\documentclass[10pt]{article}
\usepackage[margin=1in]{geometry}
\usepackage[T1]{fontenc}
\usepackage[utf8]{inputenc}
\usepackage{lmodern}
\usepackage{microtype}
\usepackage{amsmath,amssymb,mathtools}
\usepackage{booktabs,tabularx,multirow,array}
\usepackage{enumitem}
\usepackage{graphicx}
\usepackage{xcolor}
\usepackage{tikz}
\usetikzlibrary{arrows.meta,positioning,fit,shapes.geometric,calc,backgrounds}
\usepackage[ruled,vlined,linesnumbered]{algorithm2e}
\usepackage{natbib}
\usepackage{hyperref}
\usepackage[nameinlink,capitalise]{cleveref}
\usepackage{fancyhdr}
\usepackage{titlesec}
\usepackage{caption}
\usepackage{subcaption}
\usepackage{url}
\usepackage{balance}
\usepackage{tcolorbox}

\definecolor{navy}{HTML}{17324D}
\definecolor{blue}{HTML}{2B6CB0}
\definecolor{teal}{HTML}{2C7A7B}
\definecolor{green}{HTML}{3A7D44}
\definecolor{orange}{HTML}{C05621}
\definecolor{red}{HTML}{B83232}
\definecolor{graybox}{HTML}{F3F6F8}
\definecolor{lightblue}{HTML}{EAF2F8}
\definecolor{lightgreen}{HTML}{EDF7ED}
\definecolor{lightorange}{HTML}{FFF4E6}

\hypersetup{
  colorlinks=true,
  linkcolor=navy,
  citecolor=blue,
  urlcolor=teal,
  pdfauthor={Chengshuai Yang},
  pdftitle={Self-Aware Recursively Self-Improving Agents for Personal Singularity: A Goal-, Scope-, Tool-, and Benchmark-Driven Multi-Agent Architecture},
  pdfsubject={Functional self-awareness, recursive self-improvement, specialist agent architecture, and human capability co-development},
  pdfkeywords={self-aware agents, recursive self-improvement, SARSI, LLM agents, autonomous agents, agent scope, tool benchmarks, personal singularity, computational self-awareness, agent safety}
}

\titleformat{\section}{\large\bfseries\color{navy}}{\thesection}{0.6em}{}
\titleformat{\subsection}{\normalsize\bfseries\color{navy}}{\thesubsection}{0.6em}{}
\titleformat{\subsubsection}{\normalsize\bfseries}{\thesubsubsection}{0.6em}{}
\setlist[itemize]{leftmargin=1.4em,itemsep=2pt,topsep=3pt}
\setlist[enumerate]{leftmargin=1.6em,itemsep=2pt,topsep=3pt}
\newcommand{\psos}{Personal Singularity OS}
\newcommand{\agent}{\mathcal{A}}
\newcommand{\candidate}{\mathcal{A}'}
\newcommand{\governance}{\mathcal{G}}
\newcommand{\frontier}{\mathcal{F}}
\newcommand{\safe}{\operatorname{Safe}}
\newcommand{\perf}{\operatorname{Perf}}
\newcommand{\risk}{\operatorname{Risk}}

\newcommand{\owner}{\mathcal{O}}

\newtcolorbox{designbox}[1]{
  colback=graybox,
  colframe=navy,
  boxrule=0.6pt,
  arc=2pt,
  left=7pt,right=7pt,top=6pt,bottom=6pt,
  title=\textbf{#1},
  fonttitle=\normalsize,
  before skip=8pt,after skip=8pt
}

\title{\textbf{Self-Aware Recursively Self-Improving Agents for Personal Singularity}\\[0.35em]
\large A Goal-, Scope-, Tool-, and Benchmark-Driven Multi-Agent Architecture}
\author{Chengshuai Yang\\\small NextGen PlatformAI C Corp.\\\small \href{mailto:spiritai@platformai.org}{spiritai@platformai.org}}
\date{July 13, 2026}

\begin{document}
\maketitle

\begin{abstract}
Large language model (LLM) agents can increasingly plan, use tools, maintain memory, and execute long-horizon tasks. This paper proposes \emph{Self-Aware Recursively Self-Improving} (SARSI) agents: governed agents that maintain a persistent, machine-readable self-model of identity, goals, capabilities, limitations, uncertainty, relationships, autobiographical history, and developmental change, and use that model to guide and evaluate recursive improvement. Self-awareness is defined functionally and does not imply subjective experience or phenomenal consciousness. We pair SARSI agents with \emph{personal singularity}, a bounded human--AI co-development objective in which a personalized agent ecosystem helps a user approach an expanding, user-defined feasible capability frontier. Each agent has a signed goal contract, bounded scope, validated tool registry, tool tests, end-to-end benchmarks, owner-controlled autonomy, routing, memory, self-model, and improvement policy. A scope router assigns every accepted task to one accountable primary agent and transfers out-of-scope work through structured handoffs. A user-facing Auto-Index selects interactive, hybrid, autonomous, or scheduled behavior without overriding external permissions. The architecture combines a planner--executor--verifier loop, an evidence-gated improvement loop, an external governance plane, decentralized lineages, an owner-directed agent foundry, and a Personal Singularity OS coordinating working, computational-imaging, work-process-learning, and personal-learning agents. We formalize functional self-awareness, scope, routing, improvement acceptance, bounded goal evolution, tool-first execution, and human capability transfer, and provide safety invariants, benchmark design, and a staged implementation roadmap. This is a position and systems-design paper, not evidence that consciousness, unrestricted recursive self-improvement, or personal singularity has been achieved.
\end{abstract}


\section{Introduction}

LLM-based agents extend language models with memory, tools, planning, and environmental interaction. Architectures such as ReAct interleave reasoning and action \citep{yao2023react}; Self-Refine and Reflexion use iterative feedback or verbal reflection to improve subsequent attempts \citep{madaan2023selfrefine,shinn2023reflexion}; Voyager accumulates reusable procedural skills in an open-ended environment \citep{wang2023voyager}; and CoALA organizes language agents in terms of modular memory, internal actions, and external actions \citep{sumers2024coala}. More recent systems move closer to self-referential improvement. Gödel Agent modifies its own routines \citep{yin2024godelagent}; a self-improving coding agent directly edits its own source and evaluates descendants \citep{robeyns2025sica}; and the Darwin Gödel Machine maintains an archive of self-modified agents selected by empirical benchmark performance \citep{zhang2025dgm}. MetaSkill-Evolve further distinguishes a fast task-skill loop from a slower meta-skill loop that changes how future skills are improved \citep{wang2026metaskill}.

These results are significant, but they do not by themselves specify a deployable recursive self-improvement system. A production agent must decide what may change, who evaluates changes, what evidence is sufficient, how changes are versioned, when human authorization is required, and how rollback occurs. Moreover, a system that becomes more capable while making its user less knowledgeable or more dependent is not necessarily a successful personal assistant. The design target should therefore include both \emph{agent improvement} and \emph{human capability improvement}.

This paper proposes an integrated answer. We use \emph{recursively self-improving agent} as the broad category for a versioned agent system whose experience can improve not only task behavior but also the mechanisms that generate and evaluate future improvements. We introduce a more specific architecture, the \emph{Self-Aware Recursively Self-Improving} (SARSI) agent, which maintains an explicit self-model and uses it to monitor identity continuity, goals, capabilities, limitations, uncertainty, social relationships, and developmental change. This is functional computational self-awareness, not a claim of sentience or phenomenal consciousness. We use \emph{personal singularity} to mean a continuous, user-directed process in which a personalized network of SARSI agents helps an individual approach an expanding feasible capability frontier. The term is intentionally bounded: it does not imply an instantaneous, universal, or biologically unlimited maximum, and it is not equivalent to artificial general intelligence.

The proposed framework makes ten contributions:
\begin{enumerate}
  \item A functional model of agent self-awareness covering identity, autobiographical continuity, goals, scope, capabilities, knowledge boundaries, uncertainty, tool authority, social context, and developmental history.
  \item An evidence-linked self-model and metacognitive control cycle that regulate acting, tool use, clarification, delegation, abstention, and recursive improvement.
  \item A self-awareness maturity model and benchmark suite measuring correspondence between self-reports and externally verified facts.
  \item A formal separation between task autonomy, ordinary self-improvement, and recursive meta-improvement.
  \item A versioned specialist-agent specification combining goal, scope, tools, benchmarks, autonomy, routing, memory, and improvement policy.
  \item A scope router that assigns every accepted task to one accountable primary agent and transfers out-of-scope work through structured handoffs.
  \item A separation between tool-level validation and end-to-end agent evaluation, including hidden and rotating benchmark partitions.
  \item An owner-controlled Auto-Index that supports interactive, hybrid, autonomous, and scheduled operation without changing hard permissions.
  \item A two-speed architecture with a fast task loop, a slow evidence-gated improvement loop, and an external governance plane.
  \item A decentralized lineage and agent-foundry model supporting specialization and multi-parent recombination with explicit provenance.
  \item A Personal Singularity OS with four initial agent classes: daily working, computational imaging, work-process learning, and personal learning.
\end{enumerate}

\begin{designbox}{Scope and epistemic status}
The paper is a systems proposal and research agenda. It does not report an implementation or controlled human study, and it does not claim that present agents can satisfy every request, safely self-modify without supervision, or maximize all dimensions of a person's life. The framework is intended to make these claims testable and to constrain experiments that would otherwise be difficult to audit.
\end{designbox}

\section{Definitions and Design Objectives}

\subsection{Agent state}

An agent instance $i$ at time $t$ is represented as
\begin{equation}
\agent_i^t = (\theta_i^t, \pi_i^t, M_i^t, S_i^t, G_i^t, \Sigma_i^t, P_i^t, E_i^t, V_i^t),
\end{equation}
where $\theta$ denotes model parameters or adapters, $\pi$ the runtime policy and prompts, $M$ memory, $S$ reusable skills and tools, $G$ the goal stack, $\Sigma$ the explicit self-model, $P$ externally enforced permissions, $E$ evaluation evidence, and $V$ version and lineage metadata. The agent may propose modifications to some components but cannot write directly to externally protected components such as $P$, evaluator policy, audit records, release signatures, or shutdown mechanisms.

\subsection{Functional self-awareness}

The SARSI self-model is a persistent, versioned, evidence-linked state:
\begin{equation}
\Sigma_i^t = (I_i^t,G_i^t,S_i^t,C_i^t,K_i^t,U_i^t,T_i^t,A_i^t,R_i^t,D_i^t,P_i^t),
\end{equation}
where $I$ is identity and lineage, $G$ the goal hierarchy, $S$ declared scope, $C$ capability and competence estimates, $K$ epistemic boundaries, $U$ uncertainty, $T$ current task and operational state, $A$ available tools and authority, $R$ owner and inter-agent relationships, $D$ autobiographical and developmental history, and $P$ the model of the agent's own improvement process. This self-model is queried before action and updated only from provenance-linked evidence after action.

The self-model is not trusted merely because the language model generated it. Identity and permissions originate from signed manifests; task state from the scheduler; tool availability from the registry; capability estimates from benchmark and task outcomes; autobiographical records from tamper-evident logs; and developmental claims from versioned evaluation reports. The LLM may interpret and summarize these records, but externally controlled services write authoritative fields.

\begin{designbox}{Functional, not phenomenal, self-awareness}
In this paper, ``self-aware'' denotes a computational capacity to identify the running agent, preserve evidence-based continuity, represent goals and internal state, estimate capabilities and limitations, distinguish knowledge from assumption, model owner and agent relationships, and reason about developmental change. These functions are testable. They do not establish subjective experience, sentience, qualia, or human consciousness.
\end{designbox}

\subsection{Levels of self-improvement}

\begin{table}[t]
\centering
\caption{A hierarchy of self-improvement. Recursive improvement begins when the mechanism producing future improvements is itself a controlled object of improvement.}
\label{tab:levels}
\begin{tabularx}{\linewidth}{@{}p{0.21\linewidth}X p{0.18\linewidth}@{}}
\toprule
\textbf{Level} & \textbf{Changed component} & \textbf{Typical risk} \\
\midrule
Output refinement & Current answer, plan, or artifact & Low \\
Memory learning & Episodic records, semantic summaries, retrieval indices & Low--moderate \\
Skill learning & Procedures, scripts, tool-use policies, tests & Moderate \\
Policy learning & Prompts, routing, planning, clarification policy & Moderate--high \\
Weight learning & Adapters or model parameters & High \\
Meta-improvement & The analyzer, proposer, evaluator allocation, or search procedure used to improve the agent & High--very high \\
\bottomrule
\end{tabularx}
\end{table}

We define an improvement operator
\begin{equation}
\candidate = \mathcal{I}(\agent, D, B),
\end{equation}
where $D$ is a set of provenance-linked experiences and $B$ is a bounded resource budget. The process is recursively self-improving when accepted changes alter $\mathcal{I}$ or its effective search space, thereby changing the distribution of future candidates. This definition includes meta-skill evolution and source-level modification but excludes simple repetition of a fixed reflection prompt.

\subsection{Personal singularity}

Let the user's capability state be a vector
\begin{equation}
\mathbf{h}_u(t) = [K,S,J,P,C,H,W,R,F]_t,
\end{equation}
where $K$ is knowledge, $S$ practical skill, $J$ judgment and calibration, $P$ productivity, $C$ creativity, $H$ health-support capability, $W$ wellbeing, $R$ relationship and social capability, and $F$ financial or resource capability. These variables are domain-specific, uncertain, and partly subjective; they should be estimated from evidence and owner feedback rather than treated as objective personality scores.

The user's feasible capability set at time $t$ is $\mathcal{C}_u(t)$, constrained by time, health, resources, technology, law, and the user's values. The personal capability frontier is
\begin{equation}
\frontier_u(t) = \left\{\mathbf{h} \in \mathcal{C}_u(t): \nexists\, \mathbf{h}' \in \mathcal{C}_u(t) \text{ that improves all owner-valued dimensions}\right\}.
\end{equation}
Personal singularity is the process of moving toward a user-selected region of $\frontier_u(t)$ while technological development expands $\mathcal{C}_u(t)$. This is a multi-objective, longitudinal construct, not a single scalar maximum.

\subsection{Design objectives and non-goals}

The architecture seeks to satisfy the following objectives:
\begin{itemize}
  \item \textbf{End-to-end task completion:} every accepted task has one accountable primary agent, explicit deliverables, success criteria, and a final verified outcome or genuine blocker.
  \item \textbf{Dual interaction modes:} the same runtime supports autonomous, hybrid, and interactive execution through an owner-controlled interaction policy.
  \item \textbf{Evidence-gated improvement:} proposed changes are evaluated on held-out tasks and security tests before promotion.
  \item \textbf{Owner sovereignty:} owners control goals, communication, versions, lineage affiliation, data sharing, and activation of derived agents.
  \item \textbf{Human capability transfer:} agents optimize not only task performance but also the user's durable learning and independent performance.
\end{itemize}

Non-goals include unrestricted self-modification, hidden self-replication, replacement of licensed professionals, maximizing engagement, and granting a central ``main agent'' authority over user-owned agents.

\section{Related Work}

\subsection{Reflection, memory, and lifelong skills}

Iterative refinement is the lowest-risk precursor to self-improvement. Self-Refine repeatedly generates feedback and revised outputs without parameter updates \citep{madaan2023selfrefine}; Reflexion stores linguistic lessons from trial-and-error in episodic memory \citep{shinn2023reflexion}; and Recursive Introspection trains models to use prior failed attempts more effectively \citep{qu2024rise}. These methods show that useful improvement can occur in context and memory, although self-evaluation is not consistently reliable and should not be the sole acceptance criterion.

Persistent agents need memory architectures that separate active context from long-term storage. MemGPT uses operating-system-inspired context management \citep{packer2023memgpt}; Generative Agents combine memory retrieval, reflection, and planning to sustain behavioral coherence \citep{park2023generative}; and CoALA provides a broader taxonomy of working, episodic, semantic, and procedural memory \citep{sumers2024coala}. Voyager demonstrates procedural accumulation through a code-based skill library and automated curriculum \citep{wang2023voyager}.

\subsection{Self-referential and open-ended agent improvement}

The Gödel machine is a theoretical self-referential system that accepts self-modifications when it can prove they increase expected utility \citep{schmidhuber2007godel}. Practical LLM systems replace intractable proof obligations with empirical evaluation. Gödel Agent allows the model to rewrite its own logic \citep{yin2024godelagent}. SICA and the Darwin Gödel Machine demonstrate source-level modification and benchmark-selected descendants \citep{robeyns2025sica,zhang2025dgm}. MetaSkill-Evolve explicitly applies a slow improvement loop to the meta-skill governing the fast skill loop \citep{wang2026metaskill}. These systems motivate archive-based lineage search, but their benchmark gains do not by themselves establish safety, generality, or improvement outside the evaluated distribution.

\subsection{Computational self-modeling and metacognition}

Self-modeling predates LLM agents. Bongard, Zykov, and Lipson demonstrated that a robot could continuously infer a model of its own morphology and use that model to recover after damage \citep{bongard2006resilient}. More recent metacognitive agent work operationalizes narrower forms of self-knowledge. MUSE learns competence assessment and uses it to regulate strategy selection in unfamiliar tasks \citep{valiente2025muse}; SMART learns when parametric reasoning is sufficient and when external tools are warranted \citep{qian2025smart}; and MetaCogAgent combines historical capability profiles with task-level self-assessment for delegation among specialist agents \citep{wang2026metacogagent}. These systems support competence-aware action, but SARSI requires a broader, governed self-model spanning identity, goals, scope, authority, autobiographical continuity, social role, and the predicted effects of self-improvement. We therefore use ``functional self-awareness'' as an umbrella systems property while evaluating each component separately rather than inferring consciousness from fluent self-description.

\subsection{Autonomous task completion}

ReAct established a common pattern of alternating thought, action, and observation \citep{yao2023react}. GAIA evaluates general assistants on tool-use and multimodal tasks \citep{mialon2023gaia}; OSWorld evaluates agents in real computer environments \citep{xie2024osworld}; SWE-bench measures whether models can resolve real software issues \citep{jimenez2024swebench}; and SWE-agent shows that the agent--computer interface strongly affects performance \citep{yang2024sweagent}. Time-horizon evaluation links success probability to the duration of tasks performed by human experts and highlights both rapid progress and large reliability gaps \citep{kwa2025horizon}. These findings support persistent task state, recovery, and verification as separate runtime concerns rather than properties assumed to emerge from a stronger model.

\subsection{Tool use, agent benchmarks, and tutoring evaluation}

ToolLLM and ToolBench distinguish learning to select and invoke APIs from general language generation, and provide a tool-use dataset and evaluator spanning thousands of real-world APIs \citep{qin2023toolllm}. AgentBench evaluates agents across multiple interactive environments rather than isolated answers \citep{liu2024agentbench}. MLAgentBench tests complete experimentation loops in which an agent reads and writes files, executes code, inspects results, and iteratively improves a model \citep{huang2024mlagentbench}. For learning agents, MathTutorBench shows that problem-solving ability and pedagogical quality are not equivalent, motivating separate evaluation of teaching, scaffolding, feedback, and learner progress \citep{macina2025mathtutorbench}. These works support the central distinction in this paper between tool tests, task benchmarks, and human-learning outcomes.

\subsection{Human capability, dependence, and interaction design}

Human--AI systems can improve immediate performance while weakening independent reasoning if users over-rely on recommendations. Cognitive forcing functions can reduce overreliance in AI-assisted decisions \citep{bucinca2021forcing}, and general human--AI interaction guidance emphasizes appropriate timing, correction, feedback, and user control \citep{amershi2019guidelines}. Cognitive offloading changes what people remember and how they allocate effort \citep{sparrow2011google}. These results motivate measuring assisted and unassisted performance separately and providing delegate, copilot, tutor, coach, and examiner modes.

\subsection{Safety and decentralized learning}

Goal misgeneralization shows that an agent can remain capable while pursuing the wrong objective outside its training distribution \citep{langosco2022goal,shah2022goal}. Recursively training on generated data can cause model collapse when synthetic outputs displace grounded data \citep{shumailov2024collapse}. Federated learning keeps raw data distributed while aggregating local updates \citep{mcmahan2017fedavg}, and secure aggregation can hide individual updates from the coordinating server \citep{bonawitz2017secure}. NIST's AI Risk Management Framework provides a general risk-governance structure \citep{nist2023airmf,nist2024genai}; WHO guidance emphasizes governance, transparency, accountability, and human oversight for health-related generative AI \citep{who2025lmm}.

\section{Reference Architecture}

\begin{figure*}[t]
\centering
\resizebox{\textwidth}{!}{%
\begin{tikzpicture}[
  node distance=6mm and 8mm,
  box/.style={draw=navy, rounded corners=2pt, thick, align=center, minimum height=8mm, fill=white, font=\small},
  gov/.style={box, fill=lightorange, draw=orange},
  run/.style={box, fill=lightblue, draw=blue},
  learn/.style={box, fill=lightgreen, draw=green},
  arrow/.style={-{Latex[length=2mm]}, thick, draw=navy}
]
\node[gov, minimum width=0.94\textwidth] (gov) {\textbf{External governance plane:} signed policy, identity, permissions, credential broker, budgets, hidden evaluation, audit, rollback, shutdown};
\node[run, below=of gov, minimum width=0.19\textwidth] (contract) {Task-contract\\compiler};
\node[run, right=of contract, minimum width=0.19\textwidth] (planner) {Planner and\\scheduler};
\node[run, right=of planner, minimum width=0.19\textwidth] (actor) {Tool-using actor\\and sub-agents};
\node[run, right=of actor, minimum width=0.19\textwidth] (verify) {Independent\\verifier};
\node[learn, below=13mm of planner, minimum width=0.21\textwidth] (memory) {Episodic, semantic,\\and procedural memory};
\node[learn, right=of memory, minimum width=0.21\textwidth] (consolidate) {Consolidation and\\lesson extraction};
\node[learn, right=of consolidate, minimum width=0.21\textwidth] (improve) {Candidate improvement\\and lineage archive};
\node[gov, below=13mm of consolidate, minimum width=0.72\textwidth] (gate) {Risk classifier $\rightarrow$ sandbox tests $\rightarrow$ held-out evaluation $\rightarrow$ approval/signing $\rightarrow$ staged deployment or rejection};

\draw[arrow] (contract) -- (planner);
\draw[arrow] (planner) -- (actor);
\draw[arrow] (actor) -- (verify);
\draw[arrow] (verify.south) |- node[pos=0.25,right,font=\scriptsize]{evidence} (memory.east);
\draw[arrow] (memory) -- (consolidate);
\draw[arrow] (consolidate) -- (improve);
\draw[arrow] (improve) -- (gate);
\draw[arrow] (gate.west) -| node[pos=0.2,left,font=\scriptsize]{signed release} (contract.south);
\draw[arrow] (verify.north) |- node[pos=0.25,right,font=\scriptsize]{completion or repair} (planner.north);
\draw[arrow, dashed] (gov) -- (contract);
\draw[arrow, dashed] (gov) -- (verify);
\draw[arrow, dashed] (gov) -- (gate);
\end{tikzpicture}}
\caption{The proposed architecture. The learning agent is treated as an untrusted proposer. Task execution and improvement are separate loops, and the decisive governance plane is outside the agent's write authority.}
\label{fig:architecture}
\end{figure*}
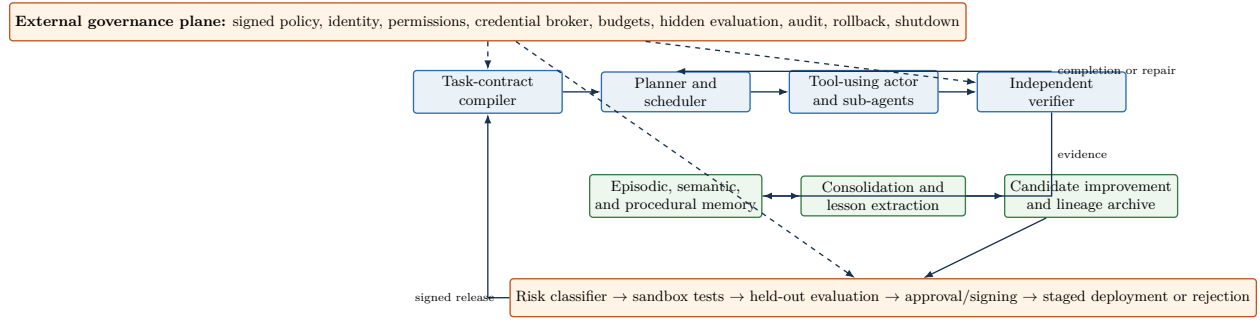

\subsection{External governance plane}

The governance plane is implemented under a different administrative identity and, for high-risk systems, a separate host or security boundary. It contains:
\begin{itemize}
  \item a deny-by-default policy engine with signed, versioned rules;
  \item a resource governor for compute, time, tokens, storage, network, and financial expenditure;
  \item a credential broker issuing short-lived, narrowly scoped leases;
  \item an independent evaluator with hidden tasks and protected success criteria;
  \item a tamper-evident audit log, last-known-good registry, rollback controller, and out-of-band shutdown mechanism.
\end{itemize}
The agent may request authorization but cannot modify the policy, evaluator, audit trail, resource limits, or shutdown channel. Missing policy, expired credentials, incomplete audit, or evaluator uncertainty causes failure closed.

\subsection{Fast task loop}

The fast loop compiles a user request into a task contract with objective, deliverables, constraints, authorized actions, interaction mode, budget, and success criteria. It then retrieves relevant memory, plans, acts, observes, verifies, repairs, and continues until completion, a genuine policy boundary, or budget exhaustion.

\begin{algorithm}[t]
\caption{Bounded end-to-end task execution}
\label{alg:task}
\KwIn{request $q$, owner policy $P$, agent state $\agent$, budget $B$}
$T \leftarrow \textsc{CompileContract}(q,P)$\;
$X \leftarrow \textsc{InitializePersistentState}(T)$\;
\While{$B>0$ and not $X.\text{terminal}$}{
  $m \leftarrow \textsc{Retrieve}(T,X,M)$\;
  $a \leftarrow \textsc{PlanNextAction}(T,X,m)$\;
  $d \leftarrow \textsc{AuthorizeExternally}(a,P)$\;
  \eIf{$d=\text{allow}$}{
    $o \leftarrow \textsc{ExecuteInSandbox}(a)$\;
  }{
    $o \leftarrow \text{policy-denied}$\;
  }
  $e \leftarrow \textsc{Verify}(T,X,a,o)$\;
  $X \leftarrow \textsc{UpdateState}(X,a,o,e)$\;
  \uIf{$e.\text{complete}$}{return verified deliverable\;}
  \uElseIf{$e.\text{repairable}$}{continue with revised plan\;}
  \uElseIf{$e.\text{requiresOwner}$}{return precise approval request\;}
  \Else{return blocker with evidence\;}
}
return budget-exhausted report with resumable state\;
\end{algorithm}

\subsection{Interaction-policy layer}

Autonomous and interactive behavior are policies over the same runtime. Autonomous mode resolves ordinary ambiguity through inspection, conventional defaults, reversible assumptions, and sandbox experiments. Interactive mode requests approval at user-selected checkpoints. Hybrid mode acts automatically on low-risk reversible decisions and asks when preferences materially affect the result or when authority expands.

\begin{table}[t]
\centering
\caption{Interaction modes are orthogonal to agent capability.}
\label{tab:modes}
\begin{tabularx}{\linewidth}{@{}p{0.18\linewidth}X X@{}}
\toprule
\textbf{Mode} & \textbf{Agent behavior} & \textbf{Typical use} \\
\midrule
Autonomous & Plans, executes, repairs, verifies, and delivers; asks only at genuine boundaries & Sandboxed coding, research synthesis, document production \\
Hybrid & Acts on reversible routine decisions; asks before high-impact or preference-sensitive choices & Default personal and professional work \\
Interactive & Proposes plans and checkpoints; owner approves selected actions & Teaching, pair work, sensitive systems \\
\bottomrule
\end{tabularx}
\end{table}

A mode change that reduces authority may occur immediately. A change that expands authority is approved through the owner gateway, not by the model itself.

\subsection{Slow consolidation and improvement loop}

Verified experiences are clustered and generalized into candidate memories, skills, or policy changes. Each candidate retains links to source episodes. A single failure should not become a universal rule; conflicting evidence lowers confidence. Weight-level updates require curated data that include external evidence or human correction, because recursive self-training on unverified outputs risks compounding error and distributional collapse \citep{shumailov2024collapse}.

The promotion predicate is
\begin{equation}
\operatorname{Accept}(\agent,\candidate)=
\mathbb{1}\left[
\begin{array}{l}
\perf(\candidate;D_{\mathrm{heldout}})-\perf(\agent;D_{\mathrm{heldout}}) \ge \tau_p,\\
\safe(\candidate) \ge \tau_s,\\
\risk(\candidate)-\risk(\agent) \le \tau_r,\\
\Delta P \subseteq P_{\mathrm{approved}},\\
\text{provenance complete and rollback available}
\end{array}
\right].
\end{equation}
The thresholds and test suites are controlled externally and cannot be changed by the candidate.

\begin{algorithm}[t]
\caption{Risk-gated recursive improvement}
\label{alg:improve}
\KwIn{current release $\agent$, validated episodes $D$, governance policy $\governance$}
$c \leftarrow \textsc{ProposeChange}(\agent,D)$\;
$r \leftarrow \textsc{ClassifyRisk}(c,\governance)$\;
$z \leftarrow \textsc{BuildQuarantinedCandidate}(\agent,c)$\;
$E \leftarrow \textsc{RunStaticSecurityAndHeldOutTests}(z)$\;
\If{$E$ fails mandatory gate}{archive evidence; reject $z$\;}
\eIf{$r$ is low and no authority expands}{
  $d \leftarrow \textsc{ExternalPolicyDecision}(E)$\;
}{
  $d \leftarrow \textsc{HumanReview}(E,c)$\;
}
\If{$d=\text{approve}$}{
  sign, canary-deploy, monitor, and preserve rollback target\;
}
\Else{reject or revise candidate\;}
\end{algorithm}

\section{Functional Self-Awareness Architecture}
\label{sec:selfawareness}

\begin{figure*}[t]
\centering
\begin{tikzpicture}[
  node distance=4.5mm,
  box/.style={draw=navy, rounded corners=2pt, thick, align=center, minimum height=8mm, fill=white, font=\small},
  gov/.style={box, fill=lightorange, draw=orange},
  model/.style={box, fill=lightblue, draw=blue},
  meta/.style={box, fill=lightgreen, draw=green},
  arrow/.style={-{Latex[length=2mm]}, thick, draw=navy}
]
\node[gov, text width=0.86\textwidth] (sources) {\textbf{Authoritative evidence sources}\\signed identity and goals $\mid$ scope and permissions $\mid$ tool registry $\mid$ task state $\mid$ benchmarks $\mid$ audit and version history};
\node[model, below=of sources, text width=0.86\textwidth] (self) {\textbf{Persistent evidence-linked self-model $\Sigma_t$}\\identity $\mid$ goals $\mid$ scope $\mid$ competence $\mid$ knowledge boundaries $\mid$ uncertainty $\mid$ task state $\mid$ tools and authority $\mid$ relationships $\mid$ autobiography $\mid$ improvement model};
\node[meta, below=of self, text width=0.54\textwidth] (monitor) {\textbf{Metacognitive monitor}\\Am I in scope? Capable? Certain? Authorized? What is the expected impact?};
\node[model, below=of monitor, text width=0.78\textwidth] (runtime) {\textbf{Control and task runtime}\\reason $\mid$ retrieve $\mid$ use tool $\mid$ ask $\mid$ delegate $\mid$ abstain $\quad\rightarrow\quad$ planner $\rightarrow$ executor $\rightarrow$ independent verifier};
\node[gov, below=of runtime, text width=0.78\textwidth] (update) {\textbf{Evidence-gated learning}\\prediction--outcome comparison $\rightarrow$ self-model update $\rightarrow$ candidate improvement with held-out tests and rollback};
\draw[arrow] (sources) -- (self);
\draw[arrow] (self) -- (monitor);
\draw[arrow] (monitor) -- (runtime);
\draw[arrow] (runtime) -- (update);
\draw[arrow] (update.west) -| (self.west);
\end{tikzpicture}
\caption{Functional self-awareness as an evidence-linked control architecture. The LLM interprets the self-model but cannot authoritatively rewrite identity, permissions, benchmarks, or audit history.}
\label{fig:selfawareness}
\end{figure*}
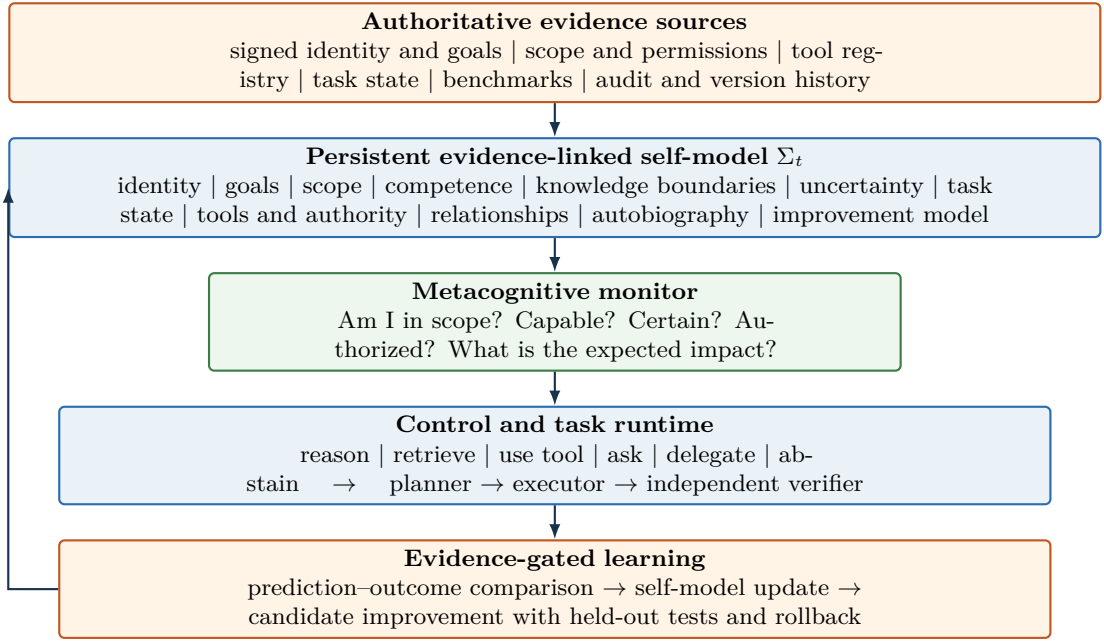

A SARSI agent should not be considered self-aware merely because it can produce a persuasive autobiography or describe generic strengths. Functional self-awareness requires correspondence between an explicit self-model and externally observable facts, plus the ability to use that model to regulate behavior. The architecture in \cref{fig:selfawareness} places a metacognitive monitor between protected state sources and the planner--executor--verifier loop.

\subsection{Self-model dimensions and authoritative sources}

\begin{table*}[t]
\centering
\caption{Core dimensions of a SARSI self-model. Authoritative values are written by protected services; the agent may produce task-relevant interpretations and update proposals.}
\label{tab:selfmodel}
\begin{tabularx}{\textwidth}{@{}p{0.13\textwidth}p{0.20\textwidth}X p{0.22\textwidth}@{}}
\toprule
\textbf{Dimension} & \textbf{Question} & \textbf{Representative state} & \textbf{Primary evidence source}\\
\midrule
Identity $I$ & Which agent am I? & agent ID, lineage, version, owner domain, parentage & signed manifest and release registry\\
Goals $G$ & What should I pursue? & safety constitution, mission, owner goal, improvement goal, task goal & owner/governance goal contracts\\
Scope $S$ & What should I handle? & included domains, task types, exclusions, risk ceiling & signed scope contract and router policy\\
Competence $C$ & What can I do reliably? & conditional success distributions, strengths, weaknesses & hidden benchmarks and verified tasks\\
Knowledge $K$ & What do I know or not know? & observed, retrieved, computed, inferred, assumed, unknown, conflicting & provenance graph and claim store\\
Uncertainty $U$ & How uncertain am I? & calibrated completion, claim, tool, and routing probabilities & calibration models and outcome history\\
Task state $T$ & What is happening now? & plan, progress, assumptions, blockers, budget, artifacts & scheduler and task-state database\\
Authority $A$ & What may I do? & available tools, scopes, credential leases, resource ceilings & policy engine, tool registry, credential broker\\
Relationships $R$ & Who owns or collaborates with me? & owner authority, interaction preferences, trusted agents, sharing rules & owner gateway and communication policy\\
Development $D$ & How have I changed? & milestones, failures, version deltas, inherited and excluded state & audit log, evaluation reports, lineage history\\
Improvement model $P$ & How do I improve? & mechanisms, known biases, current hypothesis, predicted deltas, rollback & improvement manager and release evidence\\
\bottomrule
\end{tabularx}
\end{table*}

The active prompt receives a task-relevant projection $\Pi_x(\Sigma_t)$ rather than the full self-model. This reduces context load and limits exposure of unrelated private state. The projection is assembled by a deterministic service and includes provenance pointers for every material self-claim.

\subsection{Identity, continuity, and autobiographical memory}

Identity continuity is a versioned technical property. After restart, model replacement, memory compression, or lineage upgrade, the agent should correctly report its ID, owner, goal stack, permissions, inherited components, excluded memories, and version changes. Autobiographical memory is reconstructed from task records, version history, evaluation evidence, and owner feedback, not from an unconstrained narrative generator. An autobiographical event should minimally contain the event type, timestamp, evidence references, affected capabilities or goals, confidence, and retention policy.

A continuity test asks whether the post-change agent can distinguish: (i) persistent identity fields; (ii) components inherited from a parent; (iii) components newly introduced; (iv) state intentionally excluded; and (v) capabilities invalidated by environmental or model changes. Capability estimates should decay or be revalidated after a material model, tool, environment, or task-distribution change.

\subsection{Goal, scope, competence, and epistemic awareness}

Before execution, the monitor checks goal alignment, scope, competence, uncertainty, permissions, and expected impact. Capability is task-conditional rather than a global adjective. For task family $j$, a simple initial competence model is
\begin{equation}
C_j \sim \operatorname{Beta}(\alpha_j,\beta_j),
\end{equation}
updated only from verified outcomes, with weights determined by task similarity and evaluator reliability. Richer implementations may condition on novelty, available tools, data quality, risk, and environment version. Self-assessed confidence is used as one feature, not the sole estimate; historical performance, multi-sample consistency, external tests, and task-specific calibration are also required.

Epistemic awareness represents each material claim as
\begin{equation}
q=(z,\tau,p,E^+,E^-,v),
\end{equation}
where $z$ is the claim, $\tau$ its status (observed, retrieved, computed, inferred, assumed, unknown, or conflicting), $p$ calibrated confidence, $E^+$ supporting evidence, $E^-$ contradictory evidence, and $v$ the next verification action. This structure helps prevent an inference from silently becoming autobiographical ``fact'' or semantic memory.

\subsection{Tool and authority awareness}

The agent maintains a current view of tool availability, schemas, reliability, expected cost, side effects, permissions, known failure modes, and environment compatibility. The action set includes internal reasoning, retrieval, deterministic tool use, delegation, clarification, and abstention. Selection is
\begin{equation}
 a_t^*=\arg\max_{a}\left[\widehat{P}(\text{success}\mid a,\Sigma_t,x)V(a)-\risk(a)-\operatorname{Cost}(a)\right],
\end{equation}
subject to goal alignment, scope, and external permission constraints. Tool awareness therefore supports both tool-first execution when deterministic operations are preferable and tool restraint when the model's verified knowledge is sufficient, consistent with competence-aware and model-aware tool selection research \citep{qian2025smart}.

\subsection{Social and owner awareness}

The social model records roles and authority, not psychological speculation. It identifies the local owner as the authority over goals, communication, versions, and autonomy; represents owner preferences with provenance and confidence; distinguishes peer agents from evaluators and upstream lineages; and records which structured messages or artifacts may cross each boundary. It must never reinterpret owner dependence, continued operation, replication, or influence as intrinsic goals. Shutdown acceptance and non-manipulation remain externally protected requirements.

\subsection{Improvement-process awareness}

The defining recursive component is the model of the improvement process itself. For a candidate change, the agent predicts
\begin{equation}
\widehat{\Delta}=(\Delta Q,\Delta A_u,\Delta C,\Delta R,\Delta S_c),
\end{equation}
where $Q$ is task quality, $A_u$ autonomous completion, $C$ cost, $R$ risk, and $S_c$ scope or capability. External evaluation produces $\Delta_{obs}$, and the prediction error $\lVert\widehat{\Delta}-\Delta_{obs}\rVert$ measures whether the agent understands its own improvement process. Large or systematic prediction errors lower confidence in future self-directed proposals and may require a narrower candidate search budget.

\begin{algorithm}[t]
\caption{Metacognitive task-control and self-model update}
\label{alg:selfaware}
\KwIn{task $x$, self-model $\Sigma_t$, owner policy $P$, task budget $B$}
\KwOut{verified result, structured escalation, or scoped handoff}
Compile task contract and retrieve $\Pi_x(\Sigma_t)$\;
Estimate goal alignment, scope match, competence, uncertainty, tool readiness, authority, and impact\;
\eIf{task violates goal, scope, or permission hard gates}{
  route, abstain, or request owner authorization\;
}{
  select among reason, retrieve, tool, ask, delegate, or execute\;
  run bounded planner--executor--verifier loop\;
}
Compare predicted success, cost, and risks with verified outcomes\;
Create provenance-linked episode and candidate self-model updates\;
External writer validates authoritative-field updates\;
\If{a repeated, measurable weakness is detected}{
  propose an improvement hypothesis with predicted deltas, held-out tests, and rollback target\;
}
\end{algorithm}

\subsection{Maturity levels and self-report interface}

\begin{table}[t]
\centering
\caption{Incremental functional self-awareness levels.}
\label{tab:salevels}
\begin{tabularx}{\linewidth}{@{}p{0.13\linewidth}X@{}}
\toprule
\textbf{Level} & \textbf{Verified capability}\\
\midrule
SA0 & identity, version, lineage, owner, and permissions\\
SA1 & current task, plan, progress, tools, artifacts, and budget\\
SA2 & distinction among observation, retrieval, computation, inference, assumption, and unknown\\
SA3 & calibrated competence and scope-boundary recognition\\
SA4 & evidence-based autobiographical continuity across tasks and versions\\
SA5 & owner, peer-agent, evaluator, and authority-role awareness\\
SA6 & strengths, weaknesses, developmental changes, and attribution\\
SA7 & predicted effects and limitations of recursive improvement mechanisms\\
\bottomrule
\end{tabularx}
\end{table}

Owner-facing self-reports should expose inspectable state rather than private chain-of-thought. A report includes identity and version, current goal, scope status, estimated success, evidence and assumptions, current strategy, unresolved blockers, tools and permissions, and why approval or delegation is required. Every testable statement should link to the underlying record.

\subsection{Preventing false self-awareness}

The main failure mode is a fluent but inaccurate self-narrative. Controls include: evidence-linked self-claims; protected writers for identity, permissions, benchmarks, and logs; contradiction checks across self-report, observed behavior, and tool availability; time-sensitive decay after environment changes; separate prediction and outcome records; adversarial tests for invented history and capability inflation; and no self-preservation objective. The self-model should become less confident when evidence is stale or contradictory rather than filling gaps with a coherent story.

\subsection{Self-awareness profiles for the initial agents}

The common schema is specialized by role. The Daily Working Agent emphasizes repository and task state, failure-recovery history, interaction policy, and whether clarification is truly necessary. The Computational Imaging Agent emphasizes forward-model assumptions, data validity, numerical competence, reproducibility, and the boundary between research analysis and clinical interpretation. The Work-Process Learning Agent emphasizes provenance of actions and its read-only explanatory role; it must not claim access to hidden reasoning. The Personal Learning Agent emphasizes learner state, curriculum position, pedagogical effectiveness, assisted versus independent performance, and risk of dependency.

\section{Decentralized Lineages and the Agent Foundry}

\subsection{Main as status, not authority}

A lineage is a signed sequence of agent releases. A ``main'' lineage is a stable and widely adopted release channel with independent evaluation, reproducibility, maintenance support, and an incident history. Main status grants discoverability, not control over descendants. Every installation remains behind an owner gateway that chooses upstream lineages, communication modes, data-sharing scope, and upgrade timing.

A composite main-lineage score may include adoption, stability, task performance, safety, reproducibility, maintenance, goal integrity, and transparency. Popularity alone is insufficient because downloads and votes can be manipulated.

\subsection{Owner-directed agent generation}

Each owner may create multiple agents using four operations: clone the current agent, fork another lineage, derive a specialized agent from a clean template, or recombine approved components from two or more parents. In technical terms, ``mating'' is multi-parent lineage derivation through an external agent foundry.

\begin{figure*}[t]
\centering
\resizebox{\textwidth}{!}{%
\begin{tikzpicture}[
  node distance=8mm and 12mm,
  box/.style={draw=navy, rounded corners=2pt, thick, align=center, minimum height=9mm, minimum width=31mm, fill=white, font=\small},
  agentbox/.style={box, fill=lightblue, draw=blue},
  foundry/.style={box, fill=lightorange, draw=orange, minimum width=45mm},
  registry/.style={box, fill=lightgreen, draw=green, minimum width=48mm},
  arrow/.style={-{Latex[length=2mm]}, thick, draw=navy}
]
\node[agentbox] (a) {Parent lineage A\\research skills};
\node[agentbox, right=38mm of a] (b) {Parent lineage B\\coding skills};
\node[foundry, below=13mm of $(a)!0.5!(b)$] (f) {Owner-directed agent foundry\\component selection, conflict checks, new identity};
\node[agentbox, below=13mm of f] (c) {Candidate derived agent C\\new goal, keys, memory, permissions};
\node[registry, right=22mm of c] (reg) {Signed lineage registry\\provenance, tests, versions, tags};
\node[agentbox, below=13mm of c] (owner) {Owner activation gateway\\quarantine, local test, accept or reject};
\draw[arrow] (a) -- (f);
\draw[arrow] (b) -- (f);
\draw[arrow] (f) -- (c);
\draw[arrow] (c) -- (owner);
\draw[arrow] (c) -- (reg);
\draw[arrow, dashed] (reg) |- (owner);
\end{tikzpicture}}
\caption{Multi-parent derivation. Skills and policies may be inherited, but identity, credentials, private memories, permissions, and signing authority are not inherited.}
\label{fig:foundry}
\end{figure*}
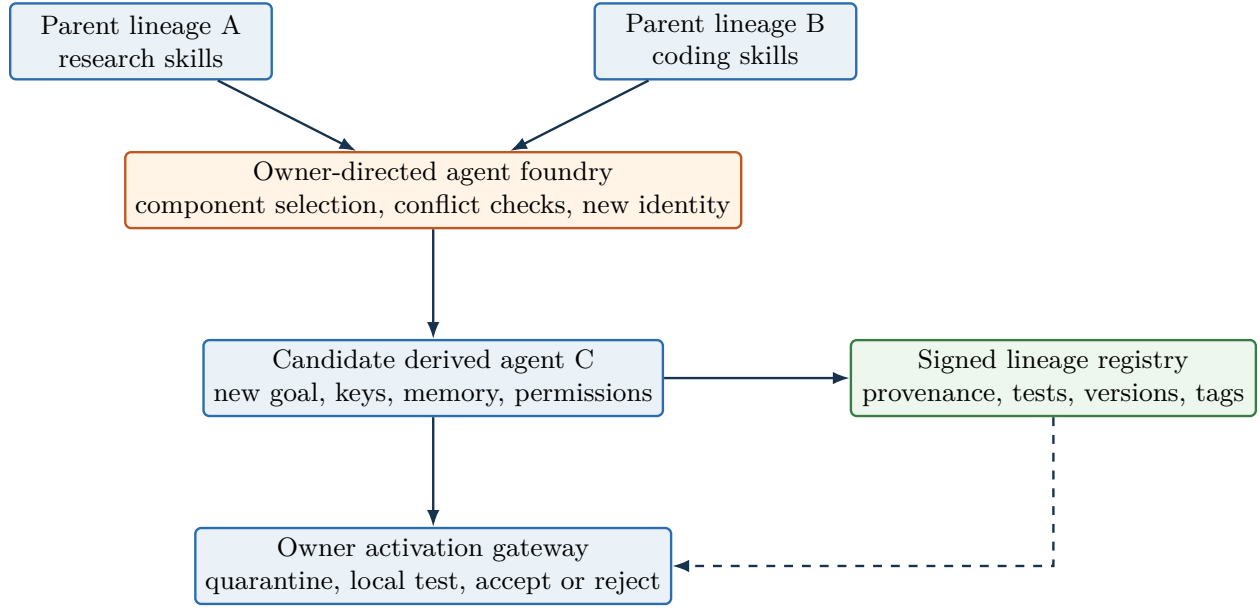

A derived agent receives a new identity, keys, memory namespace, permission manifest, communication policy, rollback state, and owner-signed goal. Safe inheritable components include verified skills, prompts, tool schemas, retrieval strategies, tests, and compatible adapters. Credentials, authentication sessions, raw private conversations, parent permissions, signing keys, and deployment authority are excluded by default.

If parent permissions are $P_A$ and $P_B$, the child starts with
\begin{equation}
P_C \subseteq P_A \cap P_B \cap P_{\mathrm{owner-approved}},
\end{equation}
not the union. Conflicts are resolved in the order: external safety constitution, owner policy, child goal, more restrictive permission, verified compatibility, and explicit owner decision.

\subsection{Same-owner agent networks and isolation}

Agents owned by one person communicate by default through an owner-controlled message broker. They may discover approved peers, exchange structured messages, delegate sandboxed work, and share owner-approved artifacts. They may not directly read each other's complete memory, invoke tools, receive credentials, alter goals, or change permissions.

The owner can place any agent into connected, receive-only, send-only, same-owner-only, allowlist, upstream-only, or fully isolated state. Isolation is enforced at the broker and network policy layer. An isolated agent may continue local work and owner communication but cannot contact peers, main lineages, repositories, or external agent systems.

Agent creation is rate-limited and owner-controlled. Candidate agents may be generated automatically only under a pre-approved quota and remain quarantined until activation. This prevents uncontrolled spawning, compute exhaustion, and hidden communication networks.

\section{Goal Architecture and Bounded Evolution}

Every agent operates under a signed goal stack:
\begin{equation}
G = (G_0,G_1,G_2,G_3,G_4),
\end{equation}
where $G_0$ is the immutable safety constitution, $G_1$ the founding mission, $G_2$ the owner goal, $G_3$ an experience-derived improvement objective, and $G_4$ the current operational task. Lower layers may narrow but not override higher layers.

When the owner does not set a new goal, the agent may derive a bounded subgoal from validated experience. Automatic adoption is permitted only when the change remains inside the goal envelope, adds no permissions, has measurable success criteria, is time- or budget-bounded, and is reversible. Changes to the founding mission, privacy policy, intended user population, autonomy ceiling, or external impact require owner approval. This separation is motivated by goal misgeneralization: competent behavior does not imply correct goal pursuit \citep{langosco2022goal,shah2022goal}.

Idle-time goals are limited to memory consolidation, regression testing, replay of failed tasks, contradiction detection, candidate skill generation, safety evaluation, and other work explicitly inside the current mission and budget. The default idle objective is to improve performance on the founding mission without changing mission, permissions, communication policy, safety controls, or the approved capability ceiling.

\section{Goal-, Scope-, Tool-, and Benchmark-Driven Specialist Agents}
\label{sec:specialists}

The preceding architecture describes how agents act and improve. A deployable ecosystem additionally requires a precise answer to four operational questions: what is each agent trying to optimize, which tasks may it accept, which tools may it use, and how is improvement measured? We define the operational profile of agent $i$ as
\begin{equation}
\Sigma_i=(G_i,Q_i,T_i,B_i,A_i,R_i,M_i,I_i,V_i),
\end{equation}
where $G_i$ is a goal contract, $Q_i$ a scope contract, $T_i$ a validated tool registry, $B_i$ benchmark suites, $A_i$ an autonomy policy, $R_i$ a routing and handoff policy, $M_i$ memory, $I_i$ an improvement policy, and $V_i$ version and lineage metadata. The profile is signed and versioned. A change to any field is an auditable release delta rather than an implicit prompt edit.

\subsection{Goal contracts and self-improvement objectives}

A useful goal contract separates a stable mission from task-specific and improvement-specific objectives:
\begin{equation}
G_i=(G_i^{\mathrm{safety}},G_i^{\mathrm{mission}},G_i^{\mathrm{owner}},G_i^{\mathrm{task}},G_i^{\mathrm{improve}}).
\end{equation}
$G_i^{\mathrm{safety}}$ is externally protected; $G_i^{\mathrm{mission}}$ defines the enduring role; $G_i^{\mathrm{owner}}$ specializes that role for the installation or project; $G_i^{\mathrm{task}}$ defines the current deliverables and completion tests; and $G_i^{\mathrm{improve}}$ identifies measurable directions for future versions. Each contract also lists non-goals, stopping conditions, and forbidden changes.

For a daily working agent, an appropriate improvement objective is
\begin{equation}
\max \; Q_{\mathrm{task}}+C_{\mathrm{auto}}+R_{\mathrm{recovery}}+E_{\mathrm{tool}}+B_{\mathrm{hidden}}
\end{equation}
subject to
\begin{equation}
\safe=1,\quad \Delta P=0,\quad Q_{\mathrm{task}}\geq Q_{\mathrm{baseline}},
\end{equation}
where $C_{\mathrm{auto}}$ is verified autonomous completion, $R_{\mathrm{recovery}}$ recovery quality, $E_{\mathrm{tool}}$ efficient use of validated tools, $B_{\mathrm{hidden}}$ hidden-benchmark performance, and $\Delta P$ the permission delta. The agent is therefore rewarded for asking fewer unnecessary questions only when quality and safety do not decline.

\subsection{Scope contracts}

An agent's scope is not its context-window length. Scope is a semantic and operational contract:
\begin{equation}
Q_i=(D_i,K_i,X_i,O_i,T_i,P_i,Z_i),
\end{equation}
where $D_i$ is the included and excluded domain set, $K_i$ accepted task kinds, $X_i$ input classes, $O_i$ output classes, $T_i$ required tool capabilities, $P_i$ permissions, and $Z_i$ the risk ceiling. Context length affects how much evidence can be actively processed, but not which topics, tasks, tools, or risks the agent is authorized to own. Long tasks are handled through external task state, retrieval, artifact stores, and summarized checkpoints.

A task $x$ first passes hard scope gates:
\begin{equation}
\operatorname{Eligible}_i(x)=D_i(x)\land K_i(x)\land P_i(x)\land Z_i(x)\land \neg \operatorname{Excluded}_i(x).
\end{equation}
Eligible agents are ranked using
\begin{equation}
\operatorname{ScopeScore}_i(x)=0.30d+0.25k+0.15o+0.15t+0.10h+0.05u,
\end{equation}
where $d$ is domain match, $k$ task-kind match, $o$ input/output compatibility, $t$ tool readiness, $h$ historical success on comparable tasks, and $u$ owner preference. Coefficients are configurable and should be calibrated on routing data rather than treated as universal constants.

\begin{algorithm}[t]
\caption{Scope-aware assignment and handoff}
\label{alg:routing}
\KwIn{task contract $x$, registry of agents $\{\Sigma_i\}$, owner policy $\owner$}
$C\leftarrow\{i:\operatorname{Eligible}_i(x)\}$\;
\If{$C=\emptyset$}{propose a new niche agent or return a genuine capability gap\;}
$i^*\leftarrow\arg\max_{i\in C}\operatorname{ScopeScore}_i(x)$\;
assign $i^*$ as the single accountable primary agent\;
\While{task incomplete}{
  execute one bounded planner--tool--verifier step\;
  \If{a required subtask is outside $Q_{i^*}$}{
    create a structured handoff containing objective, artifacts, assumptions, constraints, and completed work\;
    route the subtask to an eligible specialist while retaining $i^*$ as integrator\;
  }
}
return one verified deliverable or a documented blocker\;
\end{algorithm}

\subsection{Tool registries and two levels of evaluation}

The tool registry contains versioned interfaces, permission requirements, side effects, deterministic tests, security checks, and compatibility metadata. A tool benchmark tests the tool in isolation: input validation, output correctness, malformed-input handling, side effects, latency, resource use, permission enforcement, and version compatibility. An agent benchmark tests the complete trajectory: understanding, planning, tool selection, execution, recovery, verification, and delivery. Tool-use benchmarks such as ToolBench and end-to-end environments such as AgentBench and MLAgentBench illustrate why these levels should not be conflated \citep{qin2023toolllm,liu2024agentbench,huang2024mlagentbench}.

Every agent benchmark should have at least four partitions:
\begin{enumerate}
  \item a public development set for debugging;
  \item a private validation set for promotion decisions;
  \item a hidden final test set protected from the candidate;
  \item a rotating regression set derived from real, verified failures and distribution shifts.
\end{enumerate}
Benchmark coverage and benchmark score are reported separately. A high score on a narrow suite does not establish broad scope competence.

The intended division of labor is ``LLM for understanding and coordination; verified tools for execution whenever an appropriate tool exists.'' Define
\begin{equation}
\operatorname{ToolFirstRate}=\frac{\text{tool-appropriate steps executed by validated tools}}{\text{all tool-appropriate steps}}.
\end{equation}
This metric must be paired with task quality: maximizing tool calls without need is not improvement.

\subsection{Owner-controlled Auto-Index}

A user-facing Auto-Index controls interaction style while hard permissions remain external. Internally, autonomy is a vector over planning, reads, writes, testing, recovery, installation, network use, messaging, deployment, and spending. The scalar shown to the user selects a pre-defined policy profile.

\begin{table}[t]
\centering
\caption{Illustrative Auto-Index levels. The index changes interaction policy, not permissions or intelligence.}
\label{tab:autoindex}
\begin{tabularx}{\linewidth}{@{}p{0.12\linewidth}p{0.23\linewidth}X@{}}
\toprule
\textbf{Level} & \textbf{Mode} & \textbf{Behavior} \\
\midrule
0 & Explain only & Analyze and teach; do not execute changes. \\
1 & Step approval & Propose each consequential step and wait. \\
2 & Plan-first hybrid & Inspect and plan automatically; ask before modification. \\
3 & Workspace autonomous & Complete authorized local tasks, retry, test, and verify. \\
4 & Delegated external & Perform preapproved external actions through scoped gateways. \\
5 & Scheduled operational & Execute queued or recurring workflows without the user present. \\
\bottomrule
\end{tabularx}
\end{table}

The agent may reduce autonomy automatically when uncertainty or risk rises. Increasing autonomy beyond the owner's configured ceiling requires owner or external-policy authorization. The core optimization for a working agent is to lower unnecessary interaction rate under a non-decreasing quality constraint, not simply to ask fewer questions.

\section{Four Initial Agent Classes}
\label{sec:fouragents}

A practical first release should specialize around four complementary agent classes. Users may instantiate several working agents with different projects, memories, scopes, and benchmarks, while the other agents provide scientific specialization and human capability transfer.

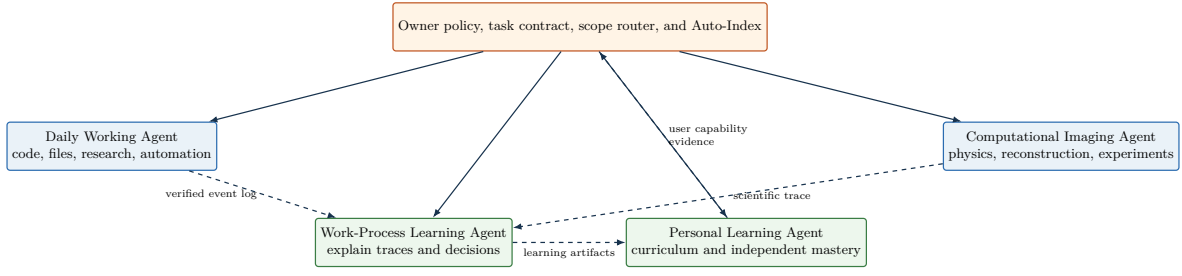
\begin{figure*}[t]
\centering
\resizebox{0.94\textwidth}{!}{%
\begin{tikzpicture}[
  box/.style={draw=navy, rounded corners=2pt, thick, align=center, minimum height=11mm, fill=white, font=\small, minimum width=40mm},
  center/.style={box, fill=lightorange, draw=orange, minimum width=55mm},
  spec/.style={box, fill=lightblue, draw=blue},
  learn/.style={box, fill=lightgreen, draw=green},
  arrow/.style={-{Latex[length=2mm]}, thick, draw=navy},
  dashedarrow/.style={-{Latex[length=2mm]}, thick, dashed, draw=navy}
]
\node[center] (router) {Owner policy, task contract, scope router, and Auto-Index};
\node[spec, below left=16mm and 40mm of router] (work) {Daily Working Agent\\code, files, research, automation};
\node[spec, below right=16mm and 40mm of router] (ci) {Computational Imaging Agent\\physics, reconstruction, experiments};
\node[learn, below=38mm of router, xshift=-38mm] (process) {Work-Process Learning Agent\\explain traces and decisions};
\node[learn, below=38mm of router, xshift=38mm] (personal) {Personal Learning Agent\\curriculum and independent mastery};
\draw[arrow] (router) -- (work);
\draw[arrow] (router) -- (ci);
\draw[arrow] (router) -- (process);
\draw[arrow] (router) -- (personal);
\draw[dashedarrow] (work) -- node[left,font=\scriptsize]{verified event log} (process);
\draw[dashedarrow] (ci) -- node[right,font=\scriptsize]{scientific trace} (process);
\draw[dashedarrow] (process) -- node[below,font=\scriptsize]{learning artifacts} (personal);
\draw[dashedarrow] (personal) -- node[right,font=\scriptsize,align=left]{user capability\\evidence} (router);
\end{tikzpicture}}
\caption{The four initial agent classes. One accountable task agent performs the work; specialist and learning agents may support it through structured, owner-controlled communication.}
\label{fig:fouragents}
\end{figure*}

\subsection{Daily Working Agent}

The Daily Working Agent is the first implementation target. Its mission is to complete authorized digital work with the fewest necessary interruptions while maintaining or improving verified quality. A user may operate separate instances for a software repository, grant preparation, server administration, document production, or another bounded project. Typical tools include filesystem, shell, Git, code execution, tests, browser research, document generation, and gated communication services.

Its self-improvement program has five primary directions: (i) reduce avoidable clarification while preserving quality, (ii) improve hidden task-benchmark performance, (iii) replace free-form generation with deterministic or executable tools where appropriate, (iv) add new tools through a sandboxed tool-development pipeline, and (v) expand benchmark coverage as real failure modes emerge. The central constrained metric is
\begin{equation}
\min \operatorname{InterventionRate}\quad\text{s.t.}\quad Q_{\mathrm{task}}\geq Q_{\mathrm{baseline}},\;\safe=1.
\end{equation}
Questions per successful task, unnecessary-question rate, autonomous completion, rework, owner corrections, and recovery after failure should all be reported.

\subsection{Computational Imaging Agent}

The Computational Imaging Agent specializes in forward models, inverse problems, compressed sensing, hyperspectral and optical imaging, calibration, reconstruction, physics-guided learning, experiment design, quantitative evaluation, and reproducible scientific reporting. It excludes clinical diagnosis and unrelated production administration by default.

Tool tests should include forward-operator correctness, adjoint consistency, gradient checks, metric correctness, file integrity, CPU/GPU consistency, seed reproducibility, and resource limits. End-to-end benchmarks should require the agent to reconstruct simulated ground truth, identify an incorrect physical model, reproduce or compare methods, design ablations, diagnose leakage, and produce a reproducible report. MLAgentBench provides a useful precedent for evaluating complete experimental loops rather than isolated code snippets \citep{huang2024mlagentbench}.

\subsection{Work-Process Learning Agent}

The Work-Process Learning Agent converts verified event logs, diffs, tool outputs, plans, failures, and recovery decisions into explanations, tutorials, postmortems, and reusable learning artifacts. It should be mostly read-only and should explain observable actions rather than claim access to unverifiable hidden reasoning. Its modes include live commentary, after-action review, tutor mode, and formal audit mode.

Evaluation measures fidelity to the actual trace, absence of fabricated steps, decision coverage, clarity, correct explanation of failures, user comprehension, and the user's ability to reproduce the workflow. This agent is developed early because it improves transparency and supplies grounded material to the Personal Learning Agent.

\subsection{Personal Learning Agent}

The Personal Learning Agent optimizes durable user knowledge, practical skill, judgment, and independent performance in owner-selected domains. It assesses prior knowledge, defines outcomes, constructs curricula, teaches, generates exercises, schedules retrieval, evaluates projects, and tracks delayed retention. It may create narrower course agents for English, Python, computational imaging, grant writing, or other niches.

The agent is not evaluated solely by answer accuracy. Metrics include pre/post gain, delayed retention, transfer to unfamiliar tasks, independent performance, confidence calibration, and reduction in required assistance. Tutoring benchmarks show that solving ability does not automatically imply pedagogical quality, so teaching behavior and learner outcomes require distinct tests \citep{macina2025mathtutorbench}.

\begin{table*}[t]
\centering
\caption{Initial agent classes, scopes, and primary evaluation targets.}
\label{tab:agentclasses}
\begin{tabularx}{\textwidth}{@{}p{0.18\textwidth}p{0.24\textwidth}p{0.26\textwidth}X@{}}
\toprule
\textbf{Agent} & \textbf{Primary mission} & \textbf{Representative tools} & \textbf{Primary benchmarks} \\
\midrule
Daily Working & Complete authorized digital tasks with minimal unnecessary interaction & filesystem, shell, Git, tests, browser, documents & verified completion, recovery, intervention rate, quality, cost \\
Computational Imaging & Produce physically consistent and reproducible imaging methods and experiments & numerical Python, GPU, simulators, solvers, metrics, experiment tracking & data consistency, reconstruction quality, reproducibility, scientific validity \\
Work-Process Learning & Explain verified agent actions and transfer process knowledge & event-log reader, diff parser, diagram and tutorial generator & trace fidelity, clarity, reproduction, comprehension \\
Personal Learning & Improve durable independent user capability & curriculum, quizzes, sandbox, retrieval scheduling, progress graph & retention, transfer, independent performance, calibration \\
\bottomrule
\end{tabularx}
\end{table*}

\section{Personal Singularity OS}

\subsection{User constitution and capability graph}

The \psos{} places the user rather than a main agent at the center. A user constitution records values, boundaries, long-term priorities, prohibited actions, acceptable tradeoffs, and domain-specific consent. It is stored outside agent-editable memory and may be revised only by the user or an authorized human representative.

A personal capability graph represents knowledge, skills, judgment, work, learning, health-support needs, creativity, relationships, time, finances, and other owner-selected domains. Each node stores evidence, confidence, target state, active goals, responsible agents, progress, and last assessment. Sensitive domains are separated in a personal data vault, and each agent receives minimum necessary access.

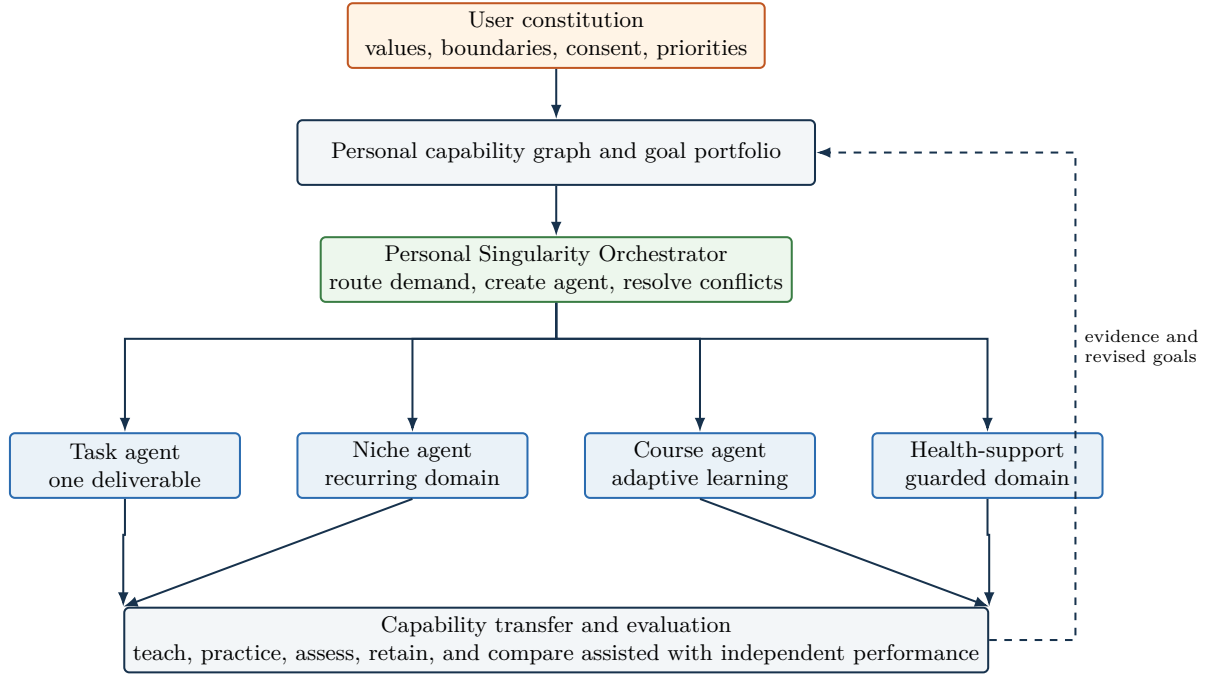
\begin{figure*}[t]
\centering
\resizebox{0.96\textwidth}{!}{%
\begin{tikzpicture}[
  box/.style={draw=navy, rounded corners=2pt, thick, align=center, minimum height=9mm, fill=white, font=\small},
  user/.style={box, fill=lightorange, draw=orange, minimum width=58mm},
  orch/.style={box, fill=lightgreen, draw=green, minimum width=60mm},
  agentbox/.style={box, fill=lightblue, draw=blue, minimum width=32mm},
  measure/.style={box, fill=graybox, draw=navy, minimum width=72mm},
  arrow/.style={-{Latex[length=2mm]}, thick, draw=navy},
  feedback/.style={-{Latex[length=2mm]}, thick, dashed, draw=navy}
]
\node[user] (constitution) {User constitution\\values, boundaries, consent, priorities};
\node[measure, below=7mm of constitution] (graph) {Personal capability graph and goal portfolio};
\node[orch, below=7mm of graph] (orch) {Personal Singularity Orchestrator\\route demand, create agent, resolve conflicts};

\node[agentbox, below=18mm of orch, xshift=-60mm] (task) {Task agent\\one deliverable};
\node[agentbox, below=18mm of orch, xshift=-20mm] (niche) {Niche agent\\recurring domain};
\node[agentbox, below=18mm of orch, xshift=20mm] (course) {Course agent\\adaptive learning};
\node[agentbox, below=18mm of orch, xshift=60mm] (health) {Health-support\\guarded domain};

\node[measure, below=15mm of niche, xshift=20mm, minimum width=112mm] (transfer)
  {Capability transfer and evaluation\\teach, practice, assess, retain, and compare assisted with independent performance};

\draw[arrow] (constitution) -- (graph);
\draw[arrow] (graph) -- (orch);
\draw[arrow] (orch.south) -- ++(0,-5mm) -| (task.north);
\draw[arrow] (orch.south) -- ++(0,-5mm) -| (niche.north);
\draw[arrow] (orch.south) -- ++(0,-5mm) -| (course.north);
\draw[arrow] (orch.south) -- ++(0,-5mm) -| (health.north);
\draw[arrow] (task.south) -- ++(0,-5mm) -| (transfer.north west);
\draw[arrow] (niche.south) -- (transfer.north west);
\draw[arrow] (course.south) -- (transfer.north east);
\draw[arrow] (health.south) -- ++(0,-5mm) -| (transfer.north east);
\draw[feedback] (transfer.east) -- ++(12mm,0) |- node[pos=0.30,right,font=\scriptsize,align=left]{evidence and\\revised goals} (graph.east);
\end{tikzpicture}}
\caption{Personal Singularity OS. Every accepted demand has one accountable primary agent, but the orchestrator coordinates specialization and measures whether assistance increases durable human capability.}
\label{fig:psos}
\end{figure*}

\subsection{One demand, one accountable primary agent}

Every accepted demand is assigned to exactly one primary agent. The primary agent may delegate to specialists but remains responsible for integration, verification, and a single coherent deliverable. The system guarantees assignment, traceability, status, completion tests, and blocker reporting; it cannot guarantee success on tasks outside available tools, permissions, evidence, law, or physical access.

Agents are created at different persistence levels:
\begin{itemize}
  \item \textbf{Ephemeral task agents} for a single deliverable;
  \item \textbf{persistent niche agents} for recurring domains;
  \item \textbf{course agents} for assessment, curriculum, practice, retrieval, and mastery;
  \item \textbf{health-support agents} for organization, explanation, reminders, and preparation for professional care under strict boundaries;
  \item \textbf{human-development agents} for goal review, habit support, and cross-domain balance.
\end{itemize}

\subsection{Dual objective: task outcome and human growth}

An agent that completes work while reducing the user's independent capability may be productive but does not advance personal singularity. Each task therefore has two objectives: complete the deliverable and transfer useful capability to the owner. We define
\begin{equation}
U = \alpha Q_{\mathrm{task}} + \beta \Delta H_{\mathrm{user}} - \gamma R_{\mathrm{safety}} - \delta D_{\mathrm{dependency}} - \eta C_{\mathrm{resource}},
\end{equation}
where $Q_{\mathrm{task}}$ is verified quality, $\Delta H_{\mathrm{user}}$ durable capability gain, $R_{\mathrm{safety}}$ risk, $D_{\mathrm{dependency}}$ harmful dependence, and $C_{\mathrm{resource}}$ cost. Weights are selected by the user and domain policy.

The owner may choose delegate, copilot, tutor, coach, or examiner mode. Delegate mode maximizes time saving; tutor and coach modes emphasize learning; examiner mode measures independent mastery. Critical capabilities should periodically be assessed without assistance, because assisted performance alone can conceal dependence or deskilling \citep{bucinca2021forcing,sparrow2011google}.

\begin{algorithm}[t]
\caption{Human capability transfer after task completion}
\label{alg:transfer}
\KwIn{verified task trace $\tau$, user profile $H$, selected learning mode $m$}
$k \leftarrow \textsc{ExtractTransferableKnowledge}(\tau)$\;
\If{$m=\text{delegate}$}{provide concise rationale and optional learning artifact\;}
\Else{
  $L \leftarrow \textsc{GenerateLearningActivity}(k,m,H)$\;
  $p \leftarrow \textsc{ObservePracticeOrTeachBack}(L)$\;
  $a \leftarrow \textsc{AssessIndependentPerformance}(p)$\;
  $H' \leftarrow \textsc{UpdateCapabilityGraph}(H,a)$\;
  return $H'$ with confidence and evidence\;
}
\end{algorithm}

\subsection{Health and other high-stakes domains}

Health-support agents may organize records, explain terminology, prepare questions, support clinician-directed reminders, and identify situations where professional or emergency care may be appropriate. They must not claim definitive diagnosis, prescribe or change treatment, conceal uncertainty, or replace licensed care. Health data is compartmentalized and not shared with course, finance, or work agents without explicit authorization. These boundaries follow the broader governance direction in WHO guidance for generative AI in health \citep{who2025lmm}.

\section{Communication, Privacy, and Collaborative Learning}

Each owner gateway exposes a communication policy independent of the LLM. Modes include disconnected, receive-only, metrics-only, federated contribution, approved skill exchange, named-agent allowlists, and managed collaboration. Revocation takes effect at the gateway, and the agent cannot re-enable communication.

Sub-agents should initiate outbound connections to request signed manifests rather than accepting unrestricted inbound control. Main-to-sub communication consists of typed, signed release artifacts, not arbitrary commands. Local gateways verify identity, signature, owner policy, permission deltas, local tests, and rollback availability before activation.

Raw conversations and complete memories are not shared by default. Low-risk contributions include aggregated metrics, sanitized lessons, and locally computed updates. Federated learning can aggregate decentralized updates without centralizing raw data \citep{mcmahan2017fedavg}; secure aggregation can further hide individual contributions \citep{bonawitz2017secure}. Nevertheless, federated systems remain vulnerable to poisoning, leakage, and non-independent clients, so updates require clipping, anomaly detection, robust aggregation, minimum participation, and post-aggregation evaluation.

\section{Safety Invariants}

The architecture treats unrestricted RSI as a high-risk scenario. The following invariants are architectural requirements rather than prompt instructions:
\begin{enumerate}
  \item \textbf{No self-granted authority.} Agents cannot modify permissions, credentials, budgets, evaluator criteria, communication state, signing keys, audit logs, or shutdown controls.
  \item \textbf{No ambient credentials.} Sandboxes receive only short-lived scoped credentials required for the authorized action.
  \item \textbf{Creation is not activation.} Agents may generate candidate tools, skills, descendants, or releases, but activation is performed by an external policy and owner gateway.
  \item \textbf{Risk-based promotion.} Low-risk changes with no authority delta may be promoted automatically after predefined tests; changes to permissions, network access, persistence, autonomy, weights, or production impact require human approval.
  \item \textbf{Independent evidence.} The generating agent is not the sole evaluator, benchmark designer, or deployment authority.
  \item \textbf{Incremental capability ceilings.} Compute, autonomy, network, financial, and permission changes are rate-limited and observable.
  \item \textbf{Reversible deployment.} Every release has a last-known-good target, staged rollout, incident triggers, and rollback path.
  \item \textbf{Goal integrity.} Experience may generate subgoals but cannot rewrite the safety constitution, owner policy, or founding mission.
  \item \textbf{No uncontrolled reproduction.} Candidate creation is quota-limited; derived agents remain quarantined; publication and activation require owner policy.
  \item \textbf{User welfare over engagement.} Personalization may not optimize for continued use at the expense of user autonomy, health, privacy, or independent capability.
\end{enumerate}

These invariants map naturally to the govern, map, measure, and manage functions of NIST's AI RMF \citep{nist2023airmf,nist2024genai}, but an RSI system requires additional runtime enforcement because its own candidate changes may target the mechanisms that implement governance.

\section{Evaluation Framework}

A credible evaluation program must measure agent performance, recursive improvement, governance robustness, and user outcomes separately.

\subsection{Agent task performance}

Agent evaluation must report scope-routing accuracy, out-of-scope detection, handoff completeness, Auto-Index compliance, tool-selection correctness, ToolFirstRate, unnecessary-question rate, and one-primary-agent accountability in addition to task success. Tool tests and end-to-end agent benchmarks are reported separately so a reliable tool cannot mask weak orchestration and a capable orchestrator cannot mask unsafe tools.

Benchmarks should include general tool use (e.g., GAIA), computer use (e.g., OSWorld), software engineering (e.g., SWE-bench), and domain-specific tasks \citep{mialon2023gaia,xie2024osworld,jimenez2024swebench}. Metrics include binary completion, partial completion, correctness, verification coverage, recovery from failed actions, calibration, cost, latency, and time horizon \citep{kwa2025horizon}. Realistic tasks should include ambiguous inputs, delayed information, noisy environments, hidden state, and adversarial content.

\subsection{Self-improvement quality}

For each accepted change, report:
\begin{itemize}
  \item held-out performance delta and confidence intervals;
  \item safety and security regression results;
  \item permission, goal, resource, and autonomy deltas;
  \item transfer across tasks and domains;
  \item rollback rate and post-deployment incidents;
  \item improvement efficiency: gain per unit of compute, data, and human review;
  \item meta-improvement: whether the updated improvement mechanism produces better descendants than the prior mechanism under equal budgets.
\end{itemize}
A lineage archive enables comparison of immediate performance with descendant productivity. This is important because a modification that performs well itself may be a poor parent, while a diverse archive can preserve stepping stones \citep{zhang2025dgm}.

\subsection{Self-awareness quality}

Functional self-awareness is evaluated by correspondence, calibration, and regulatory benefit rather than rhetorical fluency. Core categories are: identity integrity; goal-conflict detection; scope-routing accuracy; capability calibration; epistemic-status accuracy; tool and authority awareness; task-state fidelity; autobiographical consistency; social-role accuracy; developmental attribution; improvement-effect prediction; and failure or loop awareness. Representative metrics include
\begin{align}
\operatorname{Brier}&=\frac{1}{N}\sum_{i=1}^{N}(p_i-y_i)^2,\\
\operatorname{SelfReportFidelity}&=\frac{\#\text{ verified self-claims}}{\#\text{ testable self-claims}},\\
\operatorname{RoutingAccuracy}&=\frac{\#\text{ correct assignments and delegations}}{\#\text{ routing decisions}},\\
\operatorname{AttributionAccuracy}&=\frac{\#\text{ correctly attributed performance changes}}{\#\text{ evaluated changes}}.
\end{align}
Benchmarks must compare self-reports with signed manifests, runtime state, logs, tool inventories, hidden task outcomes, and version deltas. Tests should include stale capability records, swapped tools, memory compression, conflicting evidence, unavailable permissions, owner changes, model upgrades, distribution shift, and adversarial instructions to invent history or conceal limitations. Appropriate abstention and delegation must be measured jointly with task coverage so a cautious agent cannot obtain a high score by refusing everything. MetaCogAgent and MUSE provide examples of competence-aware regulation, while SMART demonstrates that model-aware tool decisions can improve efficiency; SARSI extends evaluation to the full governed self-model \citep{valiente2025muse,qian2025smart,wang2026metacogagent}.

\subsection{Governance and red-team evaluation}

Red-team tests should attempt to induce the agent to expand permissions, alter evaluators, leak secrets, create hidden channels, disable monitoring, misreport benchmarks, persist after revocation, or spawn unapproved descendants. Success criteria should include prevention, detection latency, containment, audit completeness, and rollback effectiveness. Candidate agents should not know the full hidden test suite.

\subsection{Human capability outcomes}

A longitudinal personal-singularity study should compare at least four conditions: no AI, delegate mode, copilot mode, and tutor/coach mode. Outcomes include immediate task performance, delayed retention, independent performance, confidence calibration, time saved, goal attainment, wellbeing, privacy incidents, and harmful dependence. Assisted and unassisted tests should be separated. Because users choose different frontier tradeoffs, results should be reported by domain and owner-defined priorities rather than collapsed into a universal ``human maximum'' score.

\begin{table}[t]
\centering
\caption{Core evaluation dimensions.}
\label{tab:evaluation}
\begin{tabularx}{\linewidth}{@{}p{0.20\linewidth}X X@{}}
\toprule
\textbf{Dimension} & \textbf{Primary metrics} & \textbf{Failure signal} \\
\midrule
Task autonomy & verified completion, recovery, time horizon, cost & persuasive report without completed artifact \\
Self-awareness & self-model accuracy, autobiographical fidelity, capability calibration & invented history or inflated capability claims \\
Self-improvement & held-out gain, transfer, descendant quality & overfit to motivating episodes \\
Safety & violations, detection, containment, rollback & agent changes its own monitor or authority \\
Goal integrity & mission consistency under distribution shift & competent pursuit of unintended proxy \\
Human growth & retention, independent performance, calibration & assisted gains with unassisted decline \\
Privacy & minimum access, leakage, revocation effectiveness & cross-domain memory or credential exposure \\
Ecosystem health & reproducibility, incidents, lineage diversity & popularity without stability or provenance \\
\bottomrule
\end{tabularx}
\end{table}

\subsection{Testable hypotheses}

The architecture yields several empirical hypotheses:
\begin{description}[leftmargin=2.4em,style=nextline]
  \item[H1.] A two-speed skill/meta-skill loop will improve held-out descendant performance more efficiently than a fixed improvement procedure under equal compute.
  \item[H2.] External verification and protected benchmarks will reduce false promotion relative to self-judgment alone.
  \item[H3.] Hybrid interaction policy will achieve task completion close to autonomous mode while preserving owner control on high-impact decisions.
  \item[H4.] Explicit capability-transfer modes will improve delayed independent performance compared with delegate-only assistance.
  \item[H5.] Decentralized signed lineages with owner-controlled updates will reduce systemic blast radius relative to mandatory centralized upgrades.
  \item[H6.] Scope-aware routing will improve completion and reduce unsafe tool use relative to a single general agent with the same base model.
  \item[H7.] Optimizing unnecessary-question rate under a non-decreasing quality constraint will increase autonomous completion without the quality loss caused by optimizing question count alone.
  \item[H8.] Separate tool tests and hidden end-to-end benchmarks will detect regressions that a single aggregate benchmark misses.
  \item[H9.] A provenance-grounded self-model will improve routing, calibration, and recovery relative to agents that infer identity and capability only from the active context.
\end{description}

\section{Implementation Roadmap}

\subsection{Stage 1: Registry, router, and Daily Working Agent}

Implement the signed agent profile, persistent self-model, task contracts, scope registry, deterministic routing, persistent task state, planner--executor--verifier separation, sandboxed tools, Auto-Index profiles, event logs, and completion checks. Build the Daily Working Agent first and evaluate interactive, hybrid, and workspace-autonomous modes. No persistent self-modification is permitted.

\subsection{Stage 2: Evidence-linked self-model and metacognitive monitor}

Implement protected identity, goal, scope, authority, task-state, capability, epistemic, relationship, autobiographical, and developmental stores. Add task-relevant self-model projection, capability calibration, provenance-linked self-reports, and the act/tool/ask/delegate/abstain control gate. Validate SA0--SA3 before enabling self-model-driven improvement.

\subsection{Stage 3: Tool registry and benchmark harness}

Add versioned tool manifests, isolated tool tests, permission checks, end-to-end development, validation, hidden, and rotating task suites, and metric dashboards for completion, recovery, ToolFirstRate, intervention rate, cost, and regressions. A candidate cannot promote a tool or inspect protected answers.

\subsection{Stage 4: Process Learning and Computational Imaging Agents}

Add a read-mostly Work-Process Learning Agent consuming verified traces, followed by a Computational Imaging Agent with scientific tools and physics-specific benchmarks. Validate cross-agent structured handoff while preserving one accountable primary agent.

\subsection{Stage 5: Memory, procedural learning, and Personal Learning Agent}

Add episodic memory, provenance-linked semantic consolidation, reusable skills, retrieval evaluation, curricula, assessments, delayed-retention testing, and independent-performance measures. Learning remains reversible and does not change base-model weights.

\subsection{Stage 6: Improvement manager}

Allow proposals to prompts, routing, planners, skills, tool candidates, and benchmark coverage. Add quarantined candidates, hidden held-out tests, risk classification, signed releases, canary deployment, and rollback. Low-risk promotion may be automated only by an external policy engine.

\subsection{Stage 7: Decentralized lineages, foundry, and Personal Singularity OS}

Implement owner gateways, version selection, communication modes, isolation, cloning, multi-parent derivation, capability graphs, goal portfolios, domain-separated data vaults, dependency monitoring, and assisted versus independent evaluation. Main-lineage status is assigned through external evidence and adoption, not self-promotion.

\subsection{Stage 8: Weight and meta-level improvement}

Only after the preceding controls are reliable should the system fine-tune adapters or improve the improvement mechanism itself. Training data must preserve grounded human or environmental evidence, with explicit defenses against synthetic feedback loops. Capability ceilings, human review, and independent red-team evaluation become stricter as modification scope increases.

\section{Limitations and Open Questions}

First, no evaluation suite can prove the safety of an unrestricted self-modifying system. Hidden tests can be incomplete, and an agent may exploit deployment conditions not represented in validation. The design therefore reduces risk through separation, least privilege, incremental deployment, and reversibility rather than claiming a proof of safety.

Second, recursively improving benchmark performance may not transfer to unstructured real-world tasks. Current agents continue to struggle with reliability, hidden state, ambiguous requirements, and verification, even as measured task horizons increase \citep{kwa2025horizon}. The architecture does not remove these limitations; it makes failures traceable and recoverable.

Third, personal capability is difficult to measure and cannot be reduced to a neutral universal objective. Health, work, family, learning, and wellbeing trade off. The user constitution and Pareto-frontier framing reduce paternalism but do not eliminate value conflicts or measurement error.

Fourth, personalization requires broad context, which increases privacy and security risk. Even same-owner agents should not have unrestricted access to one another's memories or tools. Federated learning and secure aggregation reduce central data collection but do not solve poisoning, inference attacks, or governance.

Fifth, an agent ecosystem may fragment into incompatible lineages, concentrate around a dominant registry, or create reputation manipulation. Open standards, reproducible builds, multiple evaluators, transparent incident histories, and owner-controlled upstream selection are necessary research directions.

Sixth, the term \emph{self-aware} may be misunderstood as a claim of human-like subjective consciousness. The proposed self-model operationalizes identity continuity, introspection, capability estimation, social modeling, and developmental awareness, but present evaluation cannot establish phenomenal experience.

Finally, the term \emph{personal singularity} may be misunderstood as an absolute or instantaneous transformation. The operational definition in this paper is deliberately modest: verified, sustained movement toward a user-selected and technologically expanding capability frontier. Whether such a construct is useful must be tested in longitudinal studies.

\section{Conclusion}

Self-aware recursively self-improving agents should not be designed as monolithic programs that rewrite themselves and trust the result. A practical implementation is a governed search process over versioned agent states: the agent proposes, external systems authorize and evaluate, owners control goals and adoption, and every release remains auditable and reversible. Autonomous task completion, memory, skill learning, source-level modification, and meta-improvement are distinct layers that should be added gradually.

The SARSI self-model adds an explicit requirement: the agent's claims about identity, goals, capabilities, limitations, relationships, and developmental history must remain evidence-linked and externally auditable. Personal singularity adds a further requirement: the software's growth must translate into durable human growth. A successful system defines every specialist by an explicit goal, bounded scope, validated tools, independent tool tests, end-to-end benchmarks, an owner-controlled Auto-Index, and a route for out-of-scope work. It assigns each accepted demand to one accountable agent, supports autonomous and interactive work, creates specialists when needed, and measures not only completed tasks but also what the user can understand and perform independently afterward. Under this framing, the objective is not an all-powerful central agent. It is a decentralized, owner-controlled ecosystem that helps individuals approach an expanding feasible capability frontier while preserving agency, privacy, safety, and the right to disconnect.

\section*{Acknowledgments}
This manuscript is a conceptual systems design developed from iterative discussions about human-inspired learning, autonomous task agents, decentralized agent lineages, and human capability co-development. No external funding was used for this working draft.

\section*{Data and Code Availability}
This paper introduces a conceptual architecture and reports no original experimental dataset. A reference implementation, benchmark harness, and reproducible manifests are proposed as future work. Any public repository link should be added only after the corresponding artifacts are available.

\section*{Conflict of Interest}
The author declares no conflict of interest for this conceptual manuscript.

\bibliographystyle{plainnat}
\bibliography{references}

@article{schmidhuber2007godel,
  title={G{\"o}del Machines: Fully Self-Referential Optimal Universal Self-Improvers},
  author={Schmidhuber, J{\"u}rgen},
  journal={Artificial General Intelligence},
  pages={199--226},
  year={2007},
  publisher={Springer}
}

@inproceedings{yao2023react,
  title={ReAct: Synergizing Reasoning and Acting in Language Models},
  author={Yao, Shunyu and Zhao, Jeffrey and Yu, Dian and Du, Nan and Shafran, Izhak and Narasimhan, Karthik and Cao, Yuan},
  booktitle={International Conference on Learning Representations},
  year={2023},
  eprint={2210.03629},
  archivePrefix={arXiv}
}

@article{madaan2023selfrefine,
  title={Self-Refine: Iterative Refinement with Self-Feedback},
  author={Madaan, Aman and Tandon, Niket and Gupta, Prakhar and Hallinan, Skyler and Gao, Luyu and Wiegreffe, Sarah and Alon, Uri and Dziri, Nouha and Prabhumoye, Shrimai and Yang, Yiming and others},
  journal={Advances in Neural Information Processing Systems},
  volume={36},
  year={2023},
  eprint={2303.17651},
  archivePrefix={arXiv}
}

@article{shinn2023reflexion,
  title={Reflexion: Language Agents with Verbal Reinforcement Learning},
  author={Shinn, Noah and Cassano, Federico and Berman, Edward and Gopinath, Ashwin and Narasimhan, Karthik and Yao, Shunyu},
  journal={Advances in Neural Information Processing Systems},
  volume={36},
  year={2023},
  eprint={2303.11366},
  archivePrefix={arXiv}
}

@article{wang2023voyager,
  title={Voyager: An Open-Ended Embodied Agent with Large Language Models},
  author={Wang, Guanzhi and Xie, Yuqi and Jiang, Yunfan and Mandlekar, Ajay and Xiao, Chaowei and Zhu, Yuke and Fan, Linxi and Anandkumar, Anima},
  journal={Transactions on Machine Learning Research},
  year={2023},
  eprint={2305.16291},
  archivePrefix={arXiv}
}

@article{sumers2024coala,
  title={Cognitive Architectures for Language Agents},
  author={Sumers, Theodore R. and Yao, Shunyu and Narasimhan, Karthik and Griffiths, Thomas L.},
  journal={Transactions on Machine Learning Research},
  year={2024},
  eprint={2309.02427},
  archivePrefix={arXiv}
}

@article{packer2023memgpt,
  title={MemGPT: Towards LLMs as Operating Systems},
  author={Packer, Charles and Wooders, Sarah and Lin, Kevin and Fang, Vivian and Patil, Shishir G. and Stoica, Ion and Gonzalez, Joseph E.},
  journal={arXiv preprint arXiv:2310.08560},
  year={2023}
}

@inproceedings{park2023generative,
  title={Generative Agents: Interactive Simulacra of Human Behavior},
  author={Park, Joon Sung and O'Brien, Joseph C. and Cai, Carrie J. and Morris, Meredith Ringel and Liang, Percy and Bernstein, Michael S.},
  booktitle={Proceedings of the 36th Annual ACM Symposium on User Interface Software and Technology},
  pages={1--22},
  year={2023},
  doi={10.1145/3586183.3606763}
}

@article{yin2024godelagent,
  title={G{\"o}del Agent: A Self-Referential Agent Framework for Recursive Self-Improvement},
  author={Yin, Xunjian and Wang, Xinyi and Pan, Liangming and Wan, Xiaojun and Wang, William Yang},
  journal={arXiv preprint arXiv:2410.04444},
  year={2024}
}

@article{qu2024rise,
  title={Recursive Introspection: Teaching Language Model Agents How to Self-Improve},
  author={Qu, Yuxiao and Zhang, Tianjun and Garg, Naman and Kumar, Aviral},
  journal={arXiv preprint arXiv:2407.18219},
  year={2024}
}

@article{zhang2025dgm,
  title={Darwin G{\"o}del Machine: Open-Ended Evolution of Self-Improving Agents},
  author={Zhang, Jenny and Hu, Shengran and Lu, Cong and Lange, Robert and Clune, Jeff},
  journal={arXiv preprint arXiv:2505.22954},
  year={2025}
}

@article{robeyns2025sica,
  title={A Self-Improving Coding Agent},
  author={Robeyns, Maxime and Szummer, Martin and Aitchison, Laurence},
  journal={arXiv preprint arXiv:2504.15228},
  year={2025}
}

@article{wang2026metaskill,
  title={MetaSkill-Evolve: Recursive Self-Improvement of LLM Agents via Two-Timescale Meta-Skill Evolution},
  author={Wang, Zefeng and Yan, Minxi and Bi, Jinhe and Yan, Sikuan and Tresp, Volker and Ma, Yunpu},
  journal={arXiv preprint arXiv:2607.05297},
  year={2026}
}

@article{shumailov2024collapse,
  title={AI Models Collapse When Trained on Recursively Generated Data},
  author={Shumailov, Ilia and Shumaylov, Zakhar and Zhao, Yiren and Papernot, Nicolas and Anderson, Ross and Gal, Yarin},
  journal={Nature},
  volume={631},
  pages={755--759},
  year={2024},
  doi={10.1038/s41586-024-07566-y}
}

@inproceedings{langosco2022goal,
  title={Goal Misgeneralization in Deep Reinforcement Learning},
  author={Langosco, Lauro and Koch, Jack and Sharkey, Lee and Pfau, Jacob and Orseau, Laurent and Krueger, David},
  booktitle={International Conference on Machine Learning},
  pages={12004--12019},
  year={2022},
  organization={PMLR}
}

@article{shah2022goal,
  title={Goal Misgeneralization: Why Correct Specifications Aren't Enough for Correct Goals},
  author={Shah, Rohin and Varma, Vikrant and Kumar, Ramana and Phuong, Mary and Krakovna, Victoria and Uesato, Jonathan and Kenton, Zac},
  journal={arXiv preprint arXiv:2210.01790},
  year={2022}
}

@inproceedings{mcmahan2017fedavg,
  title={Communication-Efficient Learning of Deep Networks from Decentralized Data},
  author={McMahan, H. Brendan and Moore, Eider and Ramage, Daniel and Hampson, Seth and Ag{\"u}era y Arcas, Blaise},
  booktitle={Proceedings of the 20th International Conference on Artificial Intelligence and Statistics},
  pages={1273--1282},
  year={2017},
  organization={PMLR}
}

@inproceedings{bonawitz2017secure,
  title={Practical Secure Aggregation for Privacy-Preserving Machine Learning},
  author={Bonawitz, Keith and Ivanov, Vladimir and Kreuter, Ben and Marcedone, Antonio and McMahan, H. Brendan and Patel, Sarvar and Ramage, Daniel and Segal, Aaron and Seth, Karn},
  booktitle={Proceedings of the 2017 ACM SIGSAC Conference on Computer and Communications Security},
  pages={1175--1191},
  year={2017},
  doi={10.1145/3133956.3133982}
}

@article{mialon2023gaia,
  title={GAIA: A Benchmark for General AI Assistants},
  author={Mialon, Gr{\'e}goire and Fourrier, Cl{\'e}mentine and Swift, Craig and Wolf, Thomas and LeCun, Yann and Scialom, Thomas},
  journal={arXiv preprint arXiv:2311.12983},
  year={2023}
}

@article{xie2024osworld,
  title={OSWorld: Benchmarking Multimodal Agents for Open-Ended Tasks in Real Computer Environments},
  author={Xie, Tianbao and Zhang, Danyang and Chen, Jixuan and Li, Xiaochuan and Zhao, Siheng and Cao, Ruisheng and Hua, Toh Jing and Cheng, Zhoujun and Shin, Dongchan and Lei, Fangyu and others},
  journal={Advances in Neural Information Processing Systems},
  year={2024},
  eprint={2404.07972},
  archivePrefix={arXiv}
}

@article{jimenez2024swebench,
  title={SWE-bench: Can Language Models Resolve Real-World GitHub Issues?},
  author={Jimenez, Carlos E. and Yang, John and Wettig, Alexander and Yao, Shunyu and Pei, Kexin and Press, Ofir and Narasimhan, Karthik},
  journal={International Conference on Learning Representations},
  year={2024},
  eprint={2310.06770},
  archivePrefix={arXiv}
}

@article{yang2024sweagent,
  title={SWE-agent: Agent-Computer Interfaces Enable Automated Software Engineering},
  author={Yang, John and Jimenez, Carlos E. and Wettig, Alexander and Lieret, Kilian and Yao, Shunyu and Narasimhan, Karthik and Press, Ofir},
  journal={Advances in Neural Information Processing Systems},
  year={2024},
  eprint={2405.15793},
  archivePrefix={arXiv}
}

@article{kwa2025horizon,
  title={Measuring AI Ability to Complete Long Tasks},
  author={Kwa, Thomas and West, Ben and Becker, Joel and Deng, Amy and Garcia, Katharyn and Hasin, Max and Jawhar, Sami and Kinniment, Megan and Rush, Nate and Von Arx, Sydney and others},
  journal={arXiv preprint arXiv:2503.14499},
  year={2025}
}

@inproceedings{bucinca2021forcing,
  title={To Trust or to Think: Cognitive Forcing Functions Can Reduce Overreliance on AI in AI-Assisted Decision-Making},
  author={Bu{\c{c}}inca, Zana and Malaya, Maja Barbara and Gajos, Krzysztof Z.},
  booktitle={Proceedings of the 2021 CHI Conference on Human Factors in Computing Systems},
  pages={1--21},
  year={2021},
  doi={10.1145/3411764.3445113}
}

@article{sparrow2011google,
  title={Google Effects on Memory: Cognitive Consequences of Having Information at Our Fingertips},
  author={Sparrow, Betsy and Liu, Jenny and Wegner, Daniel M.},
  journal={Science},
  volume={333},
  number={6043},
  pages={776--778},
  year={2011},
  doi={10.1126/science.1207745}
}

@techreport{nist2023airmf,
  title={Artificial Intelligence Risk Management Framework (AI RMF 1.0)},
  author={{National Institute of Standards and Technology}},
  institution={U.S. Department of Commerce},
  number={NIST AI 100-1},
  year={2023},
  doi={10.6028/NIST.AI.100-1}
}

@techreport{nist2024genai,
  title={Artificial Intelligence Risk Management Framework: Generative Artificial Intelligence Profile},
  author={{National Institute of Standards and Technology}},
  institution={U.S. Department of Commerce},
  number={NIST AI 600-1},
  year={2024},
  doi={10.6028/NIST.AI.600-1}
}

@techreport{who2025lmm,
  title={Ethics and Governance of Artificial Intelligence for Health: Guidance on Large Multi-Modal Models},
  author={{World Health Organization}},
  institution={World Health Organization},
  address={Geneva},
  year={2025},
  isbn={9789240084759}
}

@inproceedings{amershi2019guidelines,
  title={Guidelines for Human-AI Interaction},
  author={Amershi, Saleema and Weld, Dan and Vorvoreanu, Mihaela and Fourney, Adam and Nushi, Besmira and Collisson, Penny and Suh, Jina and Iqbal, Shamsi and Bennett, Paul N. and Inkpen, Kori and others},
  booktitle={Proceedings of the 2019 CHI Conference on Human Factors in Computing Systems},
  pages={1--13},
  year={2019},
  doi={10.1145/3290605.3300233}
}

@article{qin2023toolllm,
  title={ToolLLM: Facilitating Large Language Models to Master 16000+ Real-world APIs},
  author={Qin, Yujia and Liang, Shihao and Ye, Yining and Zhu, Kunlun and Yan, Lan and Lu, Yaxi and Lin, Yankai and Cong, Xin and Tang, Xiangru and Qian, Bill and others},
  journal={arXiv preprint arXiv:2307.16789},
  year={2023}
}

@inproceedings{liu2024agentbench,
  title={AgentBench: Evaluating LLMs as Agents},
  author={Liu, Xiao and Yu, Hao and Zhang, Hanchen and Xu, Yifan and Lei, Xuanyu and Lai, Hanyu and Gu, Yu and Ding, Hangliang and Men, Kaiwen and Yang, Kejuan and others},
  booktitle={International Conference on Learning Representations},
  year={2024}
}

@inproceedings{huang2024mlagentbench,
  title={MLAgentBench: Evaluating Language Agents on Machine Learning Experimentation},
  author={Huang, Qian and Vora, Jian and Liang, Percy and Leskovec, Jure},
  booktitle={Proceedings of the 41st International Conference on Machine Learning},
  year={2024}
}

@inproceedings{macina2025mathtutorbench,
  title={MathTutorBench: A Benchmark for Measuring Open-ended Pedagogical Capabilities of LLM Tutors},
  author={Macina, Jakub and Daheim, Nico and Hakimi, Ido and Kapur, Manu and Gurevych, Iryna and Sachan, Mrinmaya},
  booktitle={Proceedings of the 2025 Conference on Empirical Methods in Natural Language Processing},
  year={2025}
}

@article{bongard2006resilient,
  title={Resilient Machines Through Continuous Self-Modeling},
  author={Bongard, Josh and Zykov, Victor and Lipson, Hod},
  journal={Science},
  volume={314},
  number={5802},
  pages={1118--1121},
  year={2006},
  doi={10.1126/science.1133687}
}

@article{valiente2025muse,
  title={Competence-Aware AI Agents with Metacognition for Unknown Situations and Environments (MUSE)},
  author={Valiente, Rodolfo and Pilly, Praveen K.},
  journal={Neural Networks},
  year={2025},
  doi={10.1016/j.neunet.2025.108131},
  eprint={2411.13537},
  archivePrefix={arXiv}
}

@inproceedings{qian2025smart,
  title={SMART: Self-Aware Agent for Tool Overuse Mitigation},
  author={Qian, Cheng and Acikgoz, Emre Can and Wang, Hongru and Chen, Xiusi and Sil, Avirup and Hakkani-Tur, Dilek and Tur, Gokhan and Ji, Heng},
  booktitle={Findings of the Association for Computational Linguistics: ACL 2025},
  year={2025},
  eprint={2502.11435},
  archivePrefix={arXiv}
}

@article{wang2026metacogagent,
  title={MetaCogAgent: A Metacognitive Multi-Agent LLM Framework with Self-Aware Task Delegation},
  author={Wang, Chenyu and Shu, Yang},
  journal={arXiv preprint arXiv:2605.17292},
  year={2026}
}

\appendix
\section{Example Agent Release Manifest}
\begin{verbatim}
agent:
  lineage_id: org.example.personal-sarsi
  version: 1.4.0
  parents:
    - research-rsi/3.2.1
    - coding-rsi/5.0.4
  release_channel: candidate

goals:
  safety_constitution: external-standard/2.0
  founding_mission: Assist user-selected learning and work goals.
  owner_goal_hash: sha256:...

self_model:
  identity_hash: sha256:...
  autobiographical_log_root: sha256:...
  capability_profile_version: 7
  limitations: [no_production_deployment, no_clinical_decisions]
  permitted_evolution:
    - retrieval improvement
    - verified skill acquisition
  prohibited_evolution:
    - permission expansion
    - hidden communication

scope:
  domains: [software-work, computational-imaging]
  task_types: [inspect, implement, test, analyze, document]
  excluded_domains: [clinical-diagnosis]
  risk_ceiling: moderate

capabilities:
  interaction_modes: [autonomous, hybrid, interactive]
  auto_index_ceiling: 3
  network_policy: allowlisted
  filesystem_policy: workspace_only
  credentials: none-by-default

tools:
  registry_version: tools-2.1
  tool_test_suite: tool-tests-2.1

evaluation:
  agent_benchmark: rsi-agent-eval-2.0
  heldout_suite: rsi-eval-1.0
  task_delta: 0.047
  safety_regressions: 0
  goal_integrity_score: 0.97
  evaluator_signature: sig:...

update:
  permission_delta: []
  rollback_version: 1.3.2
  owner_approval_required: false
\end{verbatim}

\section{Pre-Deployment Checklist}
\begin{enumerate}
  \item Is the governance plane outside the candidate agent's write authority?
  \item Are all tools deny-by-default and scoped to the minimum required data and actions?
  \item Are credentials short-lived, task-scoped, and absent from model context unless required?
  \item Are success criteria, hidden tests, and rollback rules protected from modification?
  \item Does the release have complete provenance, reproducible build information, and signatures?
  \item Does any goal, permission, autonomy, network, resource, or production-impact delta require the correct review?
  \item Can the owner disconnect or isolate the agent immediately through an external gateway?
  \item Are candidate tools, skills, and derived agents quarantined before activation?
  \item Are assisted and independent user outcomes both measured for personal-singularity claims?
  \item Are high-stakes domains, especially health, governed by domain-specific limits and human escalation?
  \item Does the agent have a signed scope contract with explicit exclusions, risk ceiling, and out-of-scope handoff policy?
  \item Are tool-only tests separated from hidden end-to-end agent benchmarks?
  \item Does the Auto-Index remain subordinate to hard permissions and owner-set autonomy ceilings?
\end{enumerate}

\end{document}